\documentclass[10pt,twocolumn,letterpaper]{article}

\usepackage[pagenumbers]{cvpr} 

\usepackage{booktabs}
\usepackage{amssymb}
\usepackage{multirow}
\usepackage{enumitem}
\usepackage{supertabular}
\usepackage[table]{xcolor}

\definecolor{cvprblue}{rgb}{0.21,0.49,0.74}
\usepackage[pagebackref,breaklinks,colorlinks,allcolors=cvprblue]{hyperref}

\def\confName{CVPR}
\def\confYear{2027}

\title{CoTeMIC: Compositional Text Supervision for Multiple Image Clustering}

\author{
Jiaquan Wang$^{1}$ \quad
Zhuoming Li$^{1}$ \quad
Yan Lyu$^{1}$ \quad
Walid Gomaa$^{2}$ \quad
Yuheng Jia$^{1}$\thanks{Corresponding author.}\\
$^{1}$School of Computer Science and Engineering, Southeast University, China\\
$^{2}$Department of Computer Science and Engineering,\\
Egypt-Japan University of Science and Technology, Egypt
}

\begin{document}
\maketitle

\begin{abstract}
Multiple image clustering aims to discover diverse partitions of an unlabeled image set according to different semantic criteria. Without explicit supervision, existing vision-language model (VLM)-based methods obtain criterion-specific representations for clustering through simple cross-modal mapping mechanisms, which struggle to selectively capture criterion-relevant semantics from VLM image embeddings with entangled attributes. To address this limitation, we explore the potential of text as supervision beyond its conventional role as a semantic reference and propose CoTeMIC, a compositional text-supervised framework for multiple image clustering. Specifically, we construct Full-Target text pairs as compositional text supervision, enabling a text router to learn criterion-selective representation transformations. We further introduce a consistency-guided image adapter to facilitate the application of the text router to the visual domain. Finally, a criterion-specific clustering head is employed to predict cluster assignments. Extensive experiments on diverse datasets demonstrate that CoTeMIC achieves state-of-the-art performance in multiple image clustering.
\end{abstract}

\section{Introduction}
\begin{figure}[t]
    \centering
    \includegraphics[width=\linewidth]{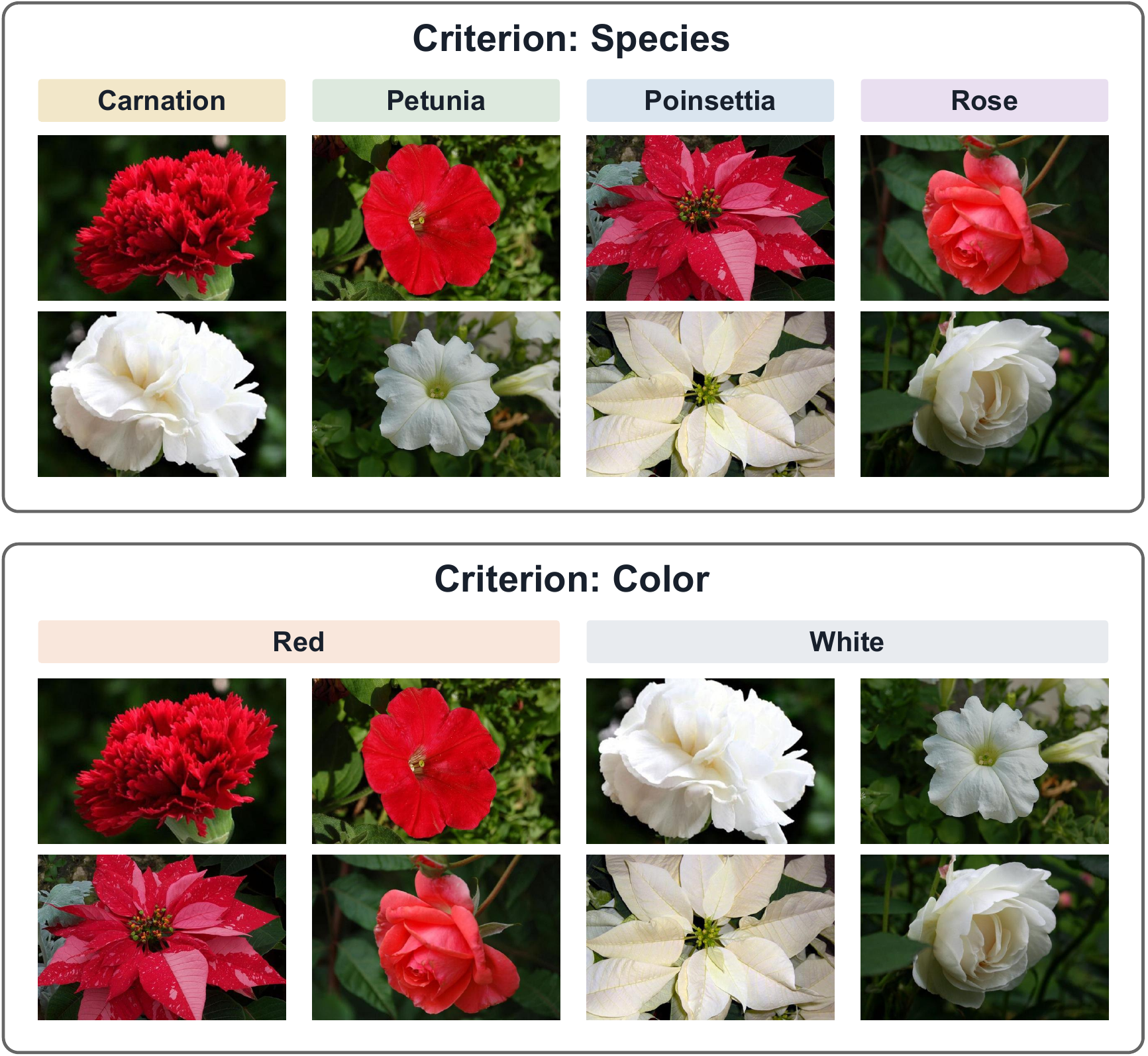}
    \caption{Illustration of multiple image clustering. The same unlabeled images can be grouped according to different semantic criteria, such as flower species and color.}
    \label{flowers}
\end{figure}

Clustering is a fundamental unsupervised learning task that aims to organize unlabeled data into meaningful groups based on their similarities. Traditional clustering methods rely on handcrafted features and predefined similarity measures to group data, whereas deep clustering methods leverage neural networks to learn discriminative representations for clustering. Despite their effectiveness, most conventional image clustering approaches focus on discovering a single partition of a dataset, which is often dominated by its most salient visual attribute. However, real-world images naturally contain multiple attributes, and the same image collection can be meaningfully grouped according to different criteria. For instance, as illustrated in Fig.~\ref{flowers}, a collection of flower images can be clustered either by species (e.g., carnation, petunia, poinsettia, and rose) or by color (e.g., red and white). This motivates multiple image clustering, which aims to discover different partitions of the same unlabeled image collection according to specified criteria.

Existing methods for multiple image clustering can be broadly categorized into three groups. Traditional methods discover different partitions through diversity-promoting strategies, such as subspace learning~\cite{hu2017finding,miklautz2020deep}, diversified attention~\cite{ren2022diversified}, data augmentation~\cite{yao2023augdmc}, and representation disentanglement~\cite{yao2024dual}. Although these methods can reveal diverse partitions, the resulting clusterings may not directly correspond to user-specified criteria. More recently, MLLM-based methods~\cite{kwon2024image, zhang2026mllm, wang2026esmc} enable clustering according to specified criteria. These methods leverage multimodal large language models (MLLMs) to extract criterion-relevant semantic information from images and perform clustering based on the extracted information. Despite their strong clustering performance, these methods can incur substantial computational overhead due to repeated querying of MLLMs, particularly on large-scale datasets.

\begin{figure}[t]
    \centering
    \includegraphics[width=\linewidth]{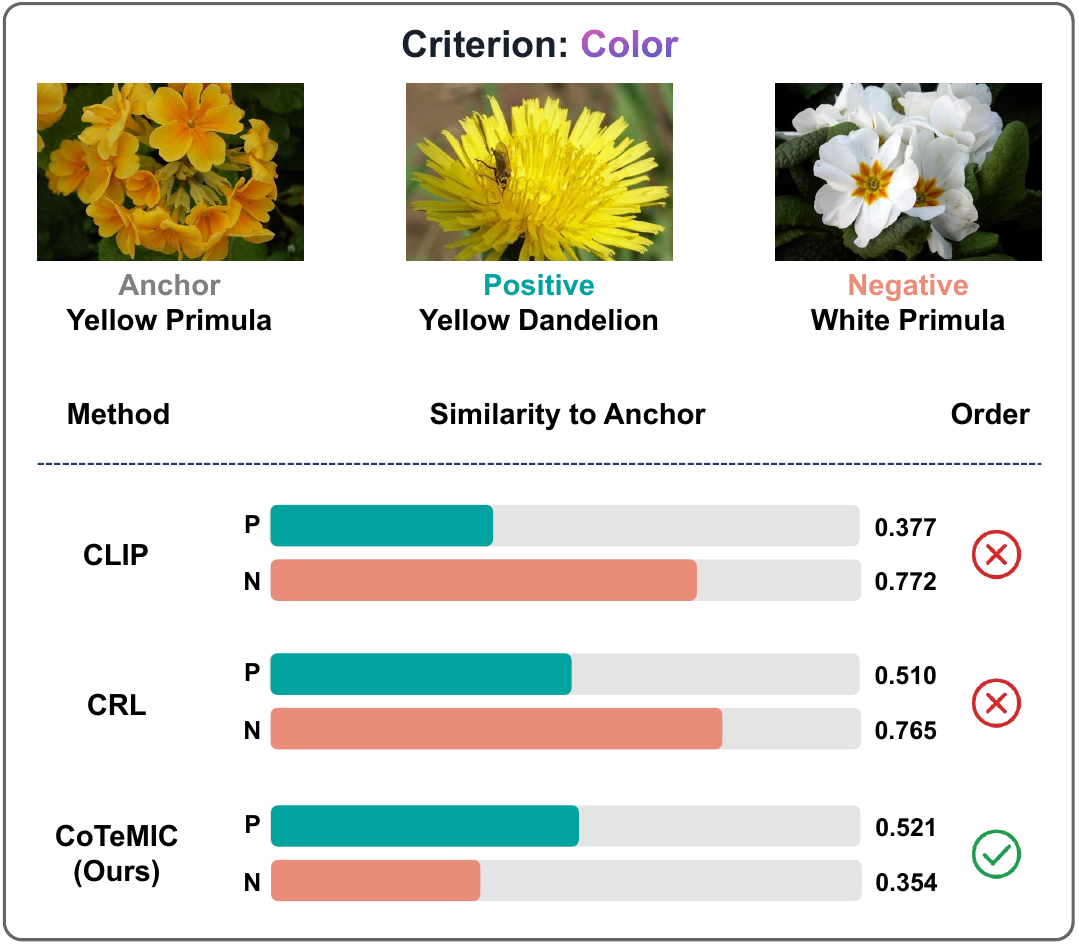}
    \caption{Illustration of semantic selectivity in similarity estimation. Under the color criterion, the yellow primula serves as the anchor, the yellow dandelion as the positive sample (same color, different species), and the white primula as the negative sample (different color, same species). CRL, like CLIP, assigns a higher similarity to the negative sample than to the positive sample. In contrast, CoTeMIC yields the desired similarity ordering, where color semantics dominate the similarity estimation, exhibiting improved criterion selectivity.}
    \label{problem}
\end{figure}

By contrast, VLM-based methods~\cite{yao2024multi, yao2024customized, liu2025conditional} leverage the image-text alignment capabilities of vision-language models (VLMs) to perform efficient multiple image clustering. Specifically, these methods employ large language models (LLMs) to generate criterion-specific textual references (e.g., red, green and blue for the color criterion), and then use them to guide the learning of criterion-specific visual representations in the VLM embedding space for subsequent clustering. In this way, they achieve criterion-aware clustering without repeatedly querying MLLMs for individual images.

However, multiple semantic attributes are often entangled in VLM image embeddings~\cite{oldfield2023parts,li2024deal}. Without explicit supervision, existing methods obtain criterion-specific representations through simple cross-modal mapping mechanisms such as linear projection or proxy learning, which struggle to selectively capture criterion-relevant semantics from these embeddings. As illustrated in Fig.~\ref{problem}, even when conditioned on color, CRL~\cite{liu2025conditional}, a representative VLM-based method, still favors species-based similarity over color-based similarity, exhibiting limited criterion selectivity. This motivates us to ask: rather than employing such limited mappings, can we directly learn criterion-selective representation transformations that emphasize criterion-relevant semantic distinctions while suppressing criterion-irrelevant semantic variations?

To this end, we explore the potential of text to provide supervision beyond its role as a semantic reference. The key observation is that natural language offers a flexible and low-cost way to compose attribute values across different criteria and isolate those relevant to a specified criterion. Specifically, by pairing descriptions that combine attribute values across different criteria with those retaining only the target attribute value, we can construct contrastive supervision that encourages representations to emphasize criterion-relevant semantic distinctions while suppressing criterion-irrelevant semantic variations. Moreover, the shared image-text embedding space of VLMs provides a foundation for extending text-based supervision to the visual domain.



Building on this insight, we propose CoTeMIC, a compositional text-supervised framework to learn criterion-specific visual representations for multiple image clustering. Specifically, we first compose full descriptions from attribute values across different criteria. Each full description is paired with a target description containing only the attribute value of the specified criterion, forming a Full–Target text pair. Using these pairs as supervision, we train a text router to learn criterion-selective representation transformations. A consistency-guided image adapter is then introduced to facilitate the transfer of text-learned transformations to the visual domain. The initial image embeddings extracted by the pretrained VLMs are adjusted by the image adapter and then passed through the text router to obtain criterion-specific visual representations. Finally, based on these representations, we train a clustering head with text-guided neighborhood consistency to predict cluster assignments. Extensive experiments on seven datasets demonstrate that CoTeMIC achieves state-of-the-art overall performance in multiple image clustering.

Our main contributions are summarized as follows:
\begin{itemize}[leftmargin=1.2em]
    \item We identify that existing VLM-based methods primarily use text as semantic references, while its potential as flexible and low-cost supervision for learning criterion-selective representation transformations remains underexplored.
    \item We propose CoTeMIC, a compositional text-supervised framework for multiple image clustering. It consists of three key components: a criterion-selective text router, a consistency-guided image adapter, and a criterion-specific clustering head.
    \item We conduct comprehensive experiments on seven datasets, demonstrating the effectiveness of CoTeMIC across multiple clustering criteria.
\end{itemize}

\section{Related Work}
\subsection{Multiple Image Clustering}
Multiple image clustering aims to discover distinct partitions of the same image collection under different semantic criteria. Traditional methods typically learn multiple non-redundant representations to uncover alternative clustering structures. MSC~\cite{hu2017finding} discovers multiple stable clusterings by maximizing Laplacian eigengaps across different feature subspaces. ENRC~\cite{miklautz2020deep} employs soft feature assignments to discover non-redundant clusterings in a shared latent space, while iMClusts~\cite{ren2022diversified} employs diversified attention to obtain multiple informative representations. AugDMC~\cite{yao2023augdmc} further exploits data augmentation to capture different aspects of visual content, and DDMC~\cite{yao2024dual} jointly learns disentangled representations and cluster assignments. Although these methods can discover diverse partitions, they offer limited control over the resulting clusterings according to user-specified criteria.

Recent advances in MLLMs have enabled criterion-aware multiple image clustering. IC$\mid$TC~\cite{kwon2024image} leverages the visual understanding and reasoning capabilities of MLLMs to group images according to the specified criterion. MLLMMC~\cite{zhang2026mllm} utilizes MLLMs for both clustering criterion discovery and criterion-specific representation learning, while ESMC~\cite{wang2026esmc} extracts criterion-relevant hidden states from MLLMs for clustering. However, processing individual images with MLLMs incurs substantial computational costs, particularly for large image collections.

Alternatively, VLM-based methods leverage pretrained image-text alignment to obtain visual representations tailored to different clustering criteria. Multi-MaP~\cite{yao2024multi} and Multi-Sub~\cite{yao2024customized} learn criterion-specific textual proxies for each image under the constraints of LLM-generated textual references. CRL~\cite{liu2025conditional} constructs conditional subspaces from LLM-generated textual references and projects image embeddings onto these subspaces to obtain criterion-specific visual representations for clustering. These methods primarily use text as semantic references to guide visual representation learning, enabling criterion-aware clustering without repeated per-image MLLM inference. However, their reliance on textual references limits their ability to suppress criterion-irrelevant semantics.

\subsection{Text Supervision for Visual Tasks}
Pretrained VLMs, such as CLIP~\cite{radford2021learning}, ALIGN~\cite{jia2021scaling}, and SigLIP~\cite{zhai2023sigmoid}, learn aligned image and text representations in a shared embedding space through large-scale image-text pretraining. This cross-modal alignment enables textual data to serve as a flexible and low-cost source of supervision for visual tasks. In particular, models or lightweight modules trained on text embeddings can be transferred to the visual domain, reducing the reliance on annotated images. However, the modality gap~\cite{liang2022mind} between image and text embeddings often requires additional adaptation for effective transfer.

Text-based supervision has been explored in various visual tasks. For multi-label image recognition, TaI-DPT~\cite{guo2023texts} learns transferable prompts from textual descriptions, while TaI-Adapter~\cite{zhu2023text} trains an adapter on text embeddings for visual recognition. CoMC~\cite{liu2024language} trains a cross-modal classifier with LLM-generated text and transfers it to image recognition through cross-modal mapping. For visual representation disentanglement, DPR~\cite{cheng2024disentangled} leverages LLM-generated descriptions to distinguish domain-invariant and domain-specific semantics, while CLAP~\cite{cai2024clap} employs text augmentation to separate content from style. In image captioning, CapDec~\cite{nukrai2022text} and DeCap~\cite{li2023decap} train language decoders using text-only data and address the image-text modality gap through noise injection and feature projection, respectively. ViECap~\cite{fei2023transferable} further introduces entity-aware prompts to improve cross-domain captioning. Text-based supervision has also been extended to MLLM training: TPI~\cite{yamabe2026text} uses rendered textual descriptions as visual inputs, while ReVision~\cite{yu2026modality} employs modality-aligned text embeddings as substitutes for image embeddings. Both reduce the reliance on paired image-text training data. Motivated by these advances, we introduce compositional text supervision to learn criterion-specific visual representations for multiple image clustering.

\begin{figure*}[t]
    \centering
    \includegraphics[width=\textwidth]{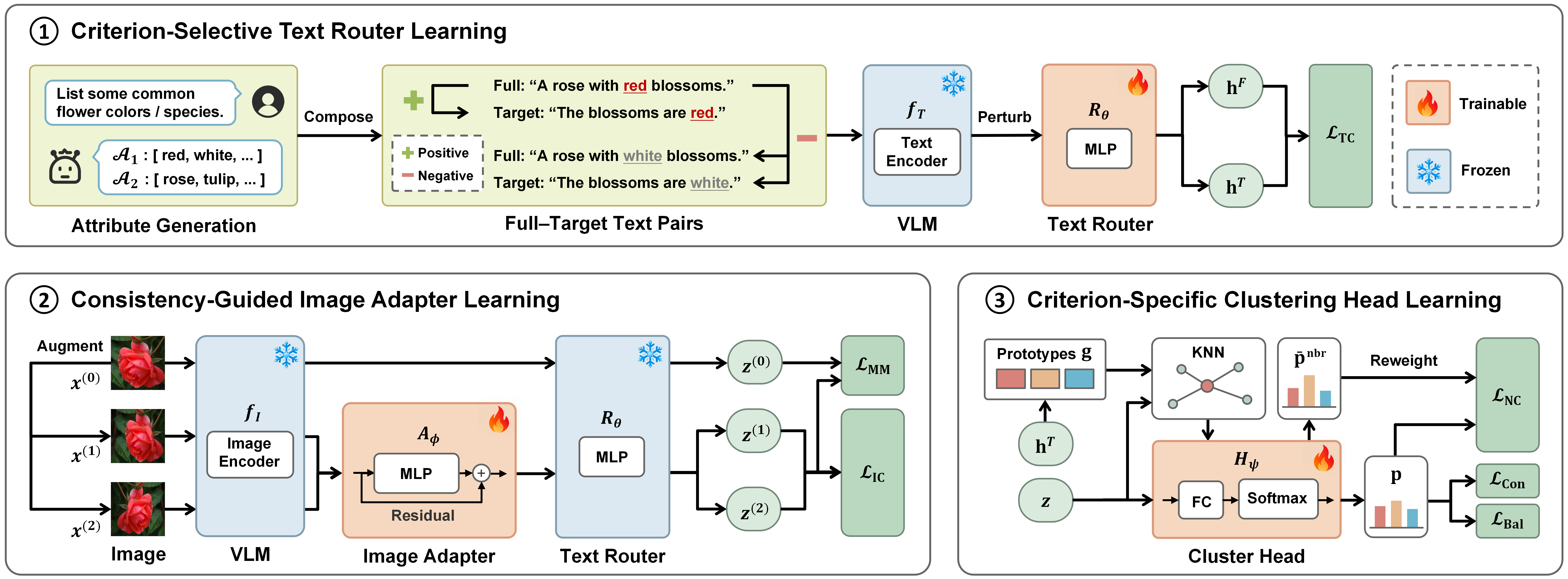}
    \caption{Overview of the proposed CoTeMIC framework, consisting of three stages: (1) compositional text-supervised router learning, (2) consistency-guided image adapter learning, and (3) criterion-specific clustering head learning.}
    \label{fig:framework}
\end{figure*}

\section{Methodology}
In this section, we present CoTeMIC, a compositional text-supervised framework for multiple image clustering, as illustrated in Fig.~\ref{fig:framework}. We first formulate the problem in Sec.~\ref{sec:problem}. Then, we construct Full-Target text pairs and use them to train a text router that performs criterion-selective transformations in Sec.~\ref{sec:router}. Next, we introduce a consistency-guided image adapter to make the learned text routers better suited for image embeddings in Sec.~\ref{sec:adapter}. Finally, we train a clustering head to obtain cluster assignments under the specified criterion in Sec.~\ref{sec:cluster}.

\subsection{Problem Formulation}
\label{sec:problem}
Given an unlabeled image collection $\mathcal{X} = \{x_i\}_{i=1}^{N}$ and its corresponding set of clustering criteria $\mathcal{C} = \{c_m\}_{m=1}^{M}$, multiple image clustering aims to partition the same image collection according to different criteria. For each criterion $c_m$, let $K_m$ denote the number of clusters, and $y_i^{(m)} \in \{1, \ldots, K_m\}$ denote the cluster assignment of image $x_i$ under criterion $c_m$.

\subsection{Criterion-Selective Text Router Learning}
\label{sec:router}
To learn criterion-selective representation transformations without image-level annotations, we train a text router using Full-Target text pairs. Specifically, following prior work~\cite{yao2024multi, yao2024customized, liu2025conditional}, we employ an LLM to generate a set of attribute values $\mathcal{A}_m$ for each criterion $c_m \in \mathcal{C}$, where $Q_m = |\mathcal{A}_m|$ denotes the number of generated attribute values. We then enumerate all possible compositions of attribute values across criteria, where each composition selects one value $a_m$ from each set $\mathcal{A}_m$.

For each composition, we construct a full description by incorporating all selected attribute values into a natural language template. Given a target criterion $c_m$, we further construct a target description using only its corresponding attribute $a_m$. Formally, the two descriptions are defined as
\begin{equation}
t^F = \mathcal{T}(a_1,\ldots,a_M), \qquad t_m^T = \mathcal{T}_m(a_m),
\end{equation}
where $\mathcal{T}$ and $\mathcal{T}_m$ denote the templates for full and target descriptions, respectively. For a given criterion (e.g., color), a Full description (e.g., ``A rose with red blossoms.'') is paired with its Target description (e.g., ``The blossoms are red.'').  More examples are provided in the supplementary material. We denote the resulting text pairs as $(t_i^F,t_i^T)$, with $\ell_i$ indicating the corresponding target attribute class.

Next, we extract text embeddings using the frozen VLM text encoder $f_T$ and center them by subtracting the mean text embedding computed over all constructed descriptions:
\begin{equation}
\mathbf{e}_i^r = f_T(t_i^r) - \boldsymbol{\mu}_{\mathrm{text}}, \qquad r \in \{F,T\}.
\end{equation}

To improve the robustness of the text router, we perturb the text embeddings with Gaussian noise before feeding them into the router:
\begin{equation}
\mathbf{h}_i^r = R_\theta\left(\mathbf{e}_i^r+\sigma\boldsymbol{\epsilon}\right), \qquad r\in\{F,T\},
\end{equation}
where $R_\theta$ denotes the text router, $\mathbf{h}_i^r$ represents the criterion-specific textual representation, $\boldsymbol{\epsilon}\sim\mathcal{N}(\mathbf{0},\mathbf{I})$ is independently sampled for each text embedding, and $\sigma$ controls the perturbation strength. Gaussian perturbation is applied only during training and omitted during inference.

To train the text router, we employ a class-masked text contrastive loss. We treat the full and target descriptions within each text pair as positives and select negatives only from different target attribute classes to avoid false negatives. For each anchor $a$, we denote the index of its paired positive by $p(a)$ and define the negative index set as
\begin{equation}
\mathcal{N}(a)=\{b\mid\ell_b\neq\ell_a\}.
\end{equation}
The text contrastive loss is then defined as
\begin{equation}
\mathcal{L}_{\mathrm{TC}}=-\frac{1}{2B_t}\sum_{a=1}^{2B_t}\log\frac{s_{a,p(a)}}{s_{a,p(a)}+\sum_{b\in\mathcal{N}(a)}s_{a,b}},
\end{equation}
where $B_t$ denotes the number of text pairs in a batch, $a$ indexes both full and target representations, $s_{a,b}=\exp(\mathrm{sim}(\mathbf{h}_a,\mathbf{h}_b)/\tau)$, $\mathrm{sim}(\cdot,\cdot)$ denotes cosine similarity, and $\tau$ is a temperature parameter. This loss emphasizes criterion-relevant semantic distinctions while suppressing criterion-irrelevant semantic variations.

\subsection{Consistency-Guided Image Adapter Learning}
\label{sec:adapter}
To facilitate the application of the text-trained router to the visual domain, we introduce a consistency-guided image adapter. Given an unlabeled image $x_i$, we define $x_i^{(0)}$ as the original view and generate two augmented views $x_i^{(1)}$ and $x_i^{(2)}$. We first extract their embeddings using the frozen VLM image encoder $f_I$ and subtract the mean image embedding:
\begin{equation}
\mathbf{v}_i^{(r)}=f_I(x_i^{(r)})-\boldsymbol{\mu}_{\mathrm{img}}, \qquad r=0,1,2.
\end{equation}
The embeddings of the two augmented views are then passed through the image adapter $A_\phi$ and the frozen text router $R_\theta$ to obtain criterion-specific visual representations:
\begin{equation}
\mathbf{z}_i^{(r)}=R_\theta\left(A_\phi(\mathbf{v}_i^{(r)})\right), \qquad r=1,2.
\end{equation}
To improve the stability of routed visual representations, we introduce an image consistency loss that minimizes the cosine distance between two augmented-view representations:
\begin{equation}
\mathcal{L}_{\mathrm{IC}}=\frac{1}{B_i}\sum_{i=1}^{B_i}\left[1-\mathrm{sim}(\mathbf{z}_i^{(1)},\mathbf{z}_i^{(2)})\right],
\end{equation}
where $B_i$ denotes the number of images in a batch, and $\mathrm{sim}(\cdot,\cdot)$ denotes cosine similarity. This objective encourages the image adapter to produce consistent criterion-specific visual representations across augmented views.

However, optimizing image consistency alone leads to undesirable distributional drift in the adapted representations. To mitigate this drift, we introduce a second-moment matching loss that aligns the second-order statistics of adapted augmented-view representations with those of original-view representations. We first apply the frozen text router directly to the original-view embeddings:
\begin{equation}
\mathbf{z}_i^{(0)}=R_\theta(\mathbf{v}_i^{(0)}).
\end{equation}
These original-view representations serve as fixed references during image adapter training. We then compute the second-moment matrices of the original-view and augmented-view representations within each batch:
\begin{equation}
\mathbf{M}^{(r)}=\frac{1}{B_i}\sum_{i=1}^{B_i}\mathbf{z}_i^{(r)}(\mathbf{z}_i^{(r)})^\top,\qquad r=0,1,2.
\end{equation}
The second-moment matching loss is defined as
\begin{equation}
\mathcal{L}_{\mathrm{MM}}=\frac{1}{2}\sum_{r=1}^{2}
\frac{\|\mathbf{M}^{(r)}-\mathbf{M}^{(0)}\|_F^2}
{\|\mathbf{M}^{(0)}\|_F^2},
\end{equation}
where $\|\cdot\|_F$ denotes the Frobenius norm.
This objective constrains batch-level representation statistics without enforcing instance-wise alignment.

The overall objective for image adapter learning is
\begin{equation}
\mathcal{L}_{\mathrm{Adapter}}=\lambda_{\mathrm{IC}}\mathcal{L}_{\mathrm{IC}}+\lambda_{\mathrm{MM}}\mathcal{L}_{\mathrm{MM}},
\end{equation}
where $\lambda_{\mathrm{IC}}$ and $\lambda_{\mathrm{MM}}$ control the contributions of the image consistency and moment matching losses, respectively.

\subsection{Criterion-Specific Clustering Head Learning}
\label{sec:cluster}
Finally, we train a clustering head  for criterion-aware image clustering. Given an image $x_i$, we obtain its criterion-specific visual representation as
\begin{equation}
\mathbf{z}_i=R_\theta(A_\phi(f_I(x_i)-\boldsymbol{\mu}_{\mathrm{img}})).
\end{equation}
The clustering head $H_\psi$ predicts soft cluster assignments for each image under the specified criterion:
\begin{equation}
\mathbf{p}_i=H_\psi(\mathbf{z}_i),
\end{equation}

To encourage confident cluster assignments, we introduce a confidence loss that minimizes the entropy of individual predictions:
\begin{equation}
\mathcal{L}_{\mathrm{Con}}=-\frac{1}{B_i}\sum_{i=1}^{B_i}\mathbf{p}_i^\top\log\mathbf{p}_i.
\end{equation}

However, minimizing prediction entropy alone may lead to degenerate solutions where most images are assigned to a few clusters. Therefore, a balance loss is introduced to prevent cluster collapse. We first compute the average cluster assignment over a batch of images:
\begin{equation}
\bar{\mathbf{p}}_b=\frac{1}{B_i}\sum_{i=1}^{B_i}\mathbf{p}_i.
\end{equation}
The balance loss is then defined as
\begin{equation}
\mathcal{L}_{\mathrm{Bal}}=\bar{\mathbf{p}}_b^\top\log\bar{\mathbf{p}}_b.
\end{equation}

\begin{table*}[ht]
\centering
\caption{Comparison of clustering performance on seven datasets under different criteria. Results are reported as percentages. Mean denotes the arithmetic average over all 16 criteria. The best and second-best available results are highlighted in \textbf{bold} and \underline{underline}, respectively.}
\label{tab:main_results}
\footnotesize
\setlength{\tabcolsep}{2.4pt}
\renewcommand{\arraystretch}{1}
\begin{tabular}{
    c
    >{\hspace{0pt}}l<{\hspace{3pt}}
    *{17}{c}
}
\toprule
\multirow{2}{*}{Metric} & \multicolumn{1}{c}{\multirow{2}{*}{Method}} & \multicolumn{4}{c}{Clevr4-10k} & \multicolumn{2}{c}{Fruit} & \multicolumn{2}{c}{Fruit360} & \multicolumn{2}{c}{CMUFace} & \multicolumn{2}{c}{Cards} & \multicolumn{2}{c}{CIFAR-10} & \multicolumn{2}{c}{Flowers} & \multirow{2}{*}{Mean} \\
\cmidrule(lr){3-6}\cmidrule(lr){7-8}\cmidrule(lr){9-10}\cmidrule(lr){11-12}\cmidrule(lr){13-14}\cmidrule(lr){15-16}\cmidrule(lr){17-18}
& & Shape & Color & Count & Texture & Species & Color & Species & Color & Pose & Glass & Number & Suit & Type & Env. & Species & Color & \\
\midrule

\multirow{15}{*}{NMI} & \multicolumn{18}{l}{\textit{Traditional Methods}} \\
& MSC & 3.60 & 58.04 & 16.58 & \underline{18.02} & 16.27 & 68.86 & 21.84 & 25.44 & 36.87 & 14.20 & 8.07 & 4.97 & 15.47 & 11.36 & 32.90 & 21.32 & 23.36 \\
& ENRC & 17.95 & 1.91 & 2.42 & 0.41 & 31.87 & 71.03 & 41.42 & 42.64 & 22.90 & 14.93 & 12.25 & 6.76 & 18.26 & 18.92 & 38.70 & 3.12 & 21.59 \\
& iMClusts & 6.40 & 1.48 & 4.49 & 1.12 & 30.29 & 73.51 & 38.61 & 40.97 & 44.37 & 19.29 & 11.44 & 7.16 & 20.40 & 19.20 & 36.54 & 39.79 & 24.69 \\
& AugDMC & 27.18 & 48.48 & 17.18 & 14.98 & 35.46 & 85.17 & 51.39 & 45.94 & 13.20 & 10.39 & 14.40 & 8.73 & 28.55 & 29.27 & 37.77 & 32.64 & 31.30 \\
& DDMC & 9.82 & 0.82 & 2.83 & 0.99 & 37.64 & \underline{89.73} & 52.92 & 49.81 & 45.26 & 22.61 & 15.63 & 9.33 & 39.91 & 37.82 & 38.22 & 50.99 & 31.52 \\
\cmidrule(lr){2-19}
& \multicolumn{18}{l}{\textit{MLLM-based Methods}} \\
& IC$\mid$TC & 31.56 & 68.48 & 27.81 & 8.22 & \textbf{100.00} & 74.79 & 29.21 & 55.88 & 16.00 & 37.67 & \textbf{61.01} & 6.48 & 59.85 & 16.02 & 41.26 & 90.61 & 45.30 \\
& MLLMMC & 61.25 & 81.11 & 27.70 & 10.11 & \textbf{100.00} & 86.70 & 46.39 & \underline{64.43} & 26.51 & 70.83 & 30.53 & 26.56 & 58.21 & 10.30 & 86.22 & \underline{91.49} & 54.90 \\
& ESMC & 72.27 & 60.39 & \underline{29.04} & 7.69 & \textbf{100.00} & 55.61 & 32.15 & 47.02 & \textbf{60.49} & 49.80 & 42.71 & \underline{35.86} & \underline{77.04} & 17.95 & 79.57 & 90.74 & 53.65 \\
\cmidrule(lr){2-19}
& \multicolumn{18}{l}{\textit{VLM-based Methods}} \\
& Multi-MaP & 54.88 & 75.17 & 21.49 & 13.66 & \textbf{100.00} & 82.16 & 50.35 & 52.05 & 29.64 & 49.72 & 32.25 & 25.42 & 49.69 & \underline{45.98} & 83.20 & 91.32 & 53.56 \\
& Multi-Sub & \underline{75.88} & 83.31 & 25.94 & 15.76 & \textbf{100.00} & 79.64 & \underline{57.66} & 55.18 & 34.59 & 67.44 & 34.83 & 32.70 & 52.71 & \textbf{48.28} & \underline{88.45} & 90.34 & 58.92 \\
& CRL & 73.11 & \underline{88.43} & 25.08 & 14.66 & \textbf{100.00} & 80.01 & 37.43 & 61.33 & 45.17 & \underline{78.89} & 32.28 & 29.44 & 72.46 & 40.82 & 87.67 & 84.40 & \underline{59.45} \\
& \cellcolor{black!10}CoTeMIC & \cellcolor{black!10}\textbf{94.98} & \cellcolor{black!10}\textbf{92.44} & \cellcolor{black!10}\textbf{30.20} & \cellcolor{black!10}\textbf{22.10} & \cellcolor{black!10}\textbf{100.00} & \cellcolor{black!10}\textbf{90.64} & \cellcolor{black!10}\textbf{69.31} & \cellcolor{black!10}\textbf{66.39} & \cellcolor{black!10}\underline{51.09} & \cellcolor{black!10}\textbf{93.08} & \cellcolor{black!10}\underline{44.28} & \cellcolor{black!10}\textbf{40.90} & \cellcolor{black!10}\textbf{80.54} & \cellcolor{black!10}40.03 & \cellcolor{black!10}\textbf{91.22} & \cellcolor{black!10}\textbf{95.16} & \cellcolor{black!10}\textbf{68.90} \\
\midrule

\multirow{15}{*}{RI} & \multicolumn{18}{l}{\textit{Traditional Methods}} \\
& MSC & 81.17 & 89.08 & 82.93 & 82.24 & 60.45 & 80.51 & 58.05 & 60.54 & 63.22 & 57.45 & 78.05 & 35.87 & 32.96 & 30.82 & 94.74 & 71.03 & 66.19 \\
& ENRC & 81.74 & 81.40 & 81.95 & 81.42 & 65.36 & 85.11 & 69.84 & 68.68 & 50.29 & 62.09 & 73.13 & 38.01 & 34.69 & 35.99 & 94.79 & 65.16 & 66.85 \\
& iMClusts & 79.88 & 79.74 & 79.81 & 79.67 & 67.43 & 86.32 & 67.32 & 68.41 & 61.14 & 56.27 & 76.58 & 37.15 & 36.95 & 36.64 & 94.76 & 75.23 & 67.71 \\
& AugDMC & 82.31 & 85.71 & 83.52 & \underline{83.07} & 73.99 & 91.08 & 74.30 & 73.92 & 55.17 & 53.61 & 82.67 & 42.28 & 45.16 & 46.89 & 94.90 & 73.17 & 71.36 \\
& DDMC & 82.36 & 81.90 & 81.97 & 81.35 & 76.21 & 93.83 & \underline{77.03} & 74.72 & 79.04 & 76.63 & 83.26 & 64.69 & 58.27 & 55.47 & 94.74 & 80.16 & 77.60 \\
\cmidrule(lr){2-19}
& \multicolumn{18}{l}{\textit{MLLM-based Methods}} \\
& IC$\mid$TC & 79.66 & 90.29 & 76.91 & 78.92 & \textbf{100.00} & 88.02 & 70.16 & 78.46 & 67.33 & 73.70 & 88.81 & 51.05 & 83.85 & 53.35 & 86.29 & 97.39 & 79.01 \\
& MLLMMC & 90.28 & 94.02 & \underline{84.89} & 80.97 & \textbf{100.00} & \underline{94.80} & 76.65 & \underline{80.77} & 65.68 & 90.25 & 87.95 & 70.22 & 82.72 & 54.63 & 98.16 & 96.80 & 84.30 \\
& ESMC & 93.24 & 89.84 & 84.80 & 82.22 & \textbf{100.00} & 74.29 & 70.72 & 73.53 & \textbf{82.64} & 77.37 & \underline{89.43} & 72.85 & \underline{92.64} & 57.49 & 97.83 & 97.55 & 83.53 \\
\cmidrule(lr){2-19}
& \multicolumn{18}{l}{\textit{VLM-based Methods}} \\
& Multi-MaP & 86.92 & 92.42 & 78.70 & 76.24 & \textbf{100.00} & 90.07 & 74.91 & 75.44 & 70.94 & 73.59 & 86.36 & 68.40 & 71.04 & 67.37 & 97.78 & \underline{97.59} & 81.74 \\
& Multi-Sub & \underline{93.74} & 94.75 & 83.79 & 82.11 & \textbf{100.00} & 91.28 & 75.69 & 77.18 & 71.78 & 88.25 & 88.02 & \underline{74.87} & 73.94 & 70.96 & \underline{98.53} & 96.40 & 85.08 \\
& CRL & 93.24 & \underline{96.71} & 83.96 & 82.85 & \textbf{100.00} & 86.79 & 69.61 & 78.74 & 76.88 & \underline{93.25} & 87.21 & 73.54 & 91.24 & \textbf{75.34} & 98.40 & 93.01 & \underline{86.30} \\
& \cellcolor{black!10}CoTeMIC & \cellcolor{black!10}\textbf{98.74} & \cellcolor{black!10}\textbf{98.13} & \cellcolor{black!10}\textbf{85.60} & \cellcolor{black!10}\textbf{84.74} & \cellcolor{black!10}\textbf{100.00} & \cellcolor{black!10}\textbf{95.90} & \cellcolor{black!10}\textbf{83.39} & \cellcolor{black!10}\textbf{83.63} & \cellcolor{black!10}\underline{80.47} & \cellcolor{black!10}\textbf{98.25} & \cellcolor{black!10}\textbf{90.80} & \cellcolor{black!10}\textbf{78.04} & \cellcolor{black!10}\textbf{94.32} & \cellcolor{black!10}\underline{74.83} & \cellcolor{black!10}\textbf{98.85} & \cellcolor{black!10}\textbf{98.70} & \cellcolor{black!10}\textbf{90.27} \\
\bottomrule
\end{tabular}
\end{table*}

To further exploit the criterion-specific semantic structure learned from text supervision, we introduce a text-guided neighborhood consistency loss. Specifically, for each target attribute class $\ell\in\{1,\ldots,Q_m\}$, we construct a text prototype by averaging the router outputs of its corresponding Target descriptions:
\begin{equation}
\mathbf{g}_\ell=\frac{1}{|\mathcal{I}_\ell|}\sum_{i\in\mathcal{I}_\ell}R_\theta(f_T(t_i^T)-\boldsymbol{\mu}_{\mathrm{text}}),
\end{equation}
where $\mathcal{I}_\ell=\{i\mid\ell_i=\ell\}$ denotes the indices of text pairs belonging to target attribute class $\ell$. For each visual representation, we compute its cosine similarities with all text prototypes to construct a text-guided response vector $\mathbf{s}_i$:
\begin{equation}
\mathbf{s}_i=\big[\mathrm{sim}(\mathbf{z}_i,\mathbf{g}_1),\ldots,\mathrm{sim}(\mathbf{z}_i,\mathbf{g}_{Q_m})\big].
\end{equation}
Based on these response vectors, we identify the $\kappa$ nearest neighbors of each image:
\begin{equation}
\mathcal{N}_{\mathrm{nbr}}(i)=\operatorname{TopK}_{j\neq i}\left(\mathrm{sim}(\mathbf{s}_i,\mathbf{s}_j);\kappa\right).
\end{equation}
The neighborhood structure is constructed once before clustering head training and remains fixed. At the beginning of each epoch, we compute the average predictions over each image's neighbors and the entire dataset:
\begin{equation}
\bar{\mathbf{p}}_i^{\mathrm{nbr}}=\frac{1}{\kappa}\sum_{j\in\mathcal{N}_{\mathrm{nbr}}(i)}\mathbf{p}_j,\qquad
\bar{\mathbf{p}}_g=\frac{1}{N}\sum_{i=1}^{N}\mathbf{p}_i.
\end{equation}
To mitigate the self-reinforcing bias toward frequently assigned clusters, we reweight the average neighbor predictions by the inverse global cluster frequencies and renormalize them to obtain neighbor-guided targets:
\begin{equation}
\mathbf{q}_i=\frac{\bar{\mathbf{p}}_i^{\mathrm{nbr}}\oslash\bar{\mathbf{p}}_g}{\left\|\bar{\mathbf{p}}_i^{\mathrm{nbr}}\oslash\bar{\mathbf{p}}_g\right\|_1}.
\end{equation}
where $\oslash$ denotes element-wise division and $\|\cdot\|_1$ denotes the $L_1$ norm. These targets are fixed during batch-wise optimization within each epoch. The neighborhood consistency loss is then defined as
\begin{equation}
\mathcal{L}_{\mathrm{NC}}=\frac{1}{B_i}\sum_{i=1}^{B_i}\|\mathbf{p}_i-\mathbf{q}_i\|_2^2.
\end{equation}

The overall objective for clustering head learning is
\begin{equation}
\mathcal{L}_{\mathrm{Head}}=\lambda_{\mathrm{Con}}\mathcal{L}_{\mathrm{Con}}+\lambda_{\mathrm{Bal}}\mathcal{L}_{\mathrm{Bal}}+\lambda_{\mathrm{NC}}\mathcal{L}_{\mathrm{NC}},
\end{equation}
where $\lambda_{\mathrm{Con}}$, $\lambda_{\mathrm{Bal}}$, and $\lambda_{\mathrm{NC}}$ are the weighting coefficients.

\begin{table*}[ht]
\centering
\footnotesize
\renewcommand{\arraystretch}{1}
\setlength{\tabcolsep}{2.4pt}
\caption{Ablation study of CoTeMIC on three datasets. Clustering accuracy (ACC, \%) is reported for each criterion. The Mean column reports the average ACC across eight criteria. The best and second-best available results are highlighted in \textbf{bold} and \underline{underline}, respectively.}
\label{tab:ablation}
\begin{tabular*}{\textwidth}{@{\extracolsep{\fill}}*{15}{c}@{}}
\toprule
\multicolumn{1}{c}{Text Router} & \multicolumn{2}{c}{Image Adapter} & \multicolumn{3}{c}{Clustering Head} & \multicolumn{4}{c}{Clevr4-10k} & \multicolumn{2}{c}{Fruit360} & \multicolumn{2}{c}{Cards} & \multirow{2}{*}{Mean} \\
\cmidrule(lr){1-1}\cmidrule(lr){2-3}\cmidrule(lr){4-6}\cmidrule(lr){7-10}\cmidrule(lr){11-12}\cmidrule(lr){13-14}
$\mathcal{L}_{\mathrm{TC}}$ & $\mathcal{L}_{\mathrm{IC}}$ & $\mathcal{L}_{\mathrm{MM}}$ & $\mathcal{L}_{\mathrm{Con}}$ & $\mathcal{L}_{\mathrm{Bal}}$ & $\mathcal{L}_{\mathrm{NC}}$ & Shape & Color & Count & Texture & Species & Color & Number & Suit & \\
\midrule
-- & -- & -- & -- & -- & -- & 77.95 & 11.96 & 19.11 & 12.97 & 66.58 & 49.11 & 18.50 & 44.55 & 37.59 \\
$\checkmark$ & -- & -- & -- & -- & -- & 85.39 & 90.46 & 32.78 & 31.33 & 68.05 & 75.86 & 51.16 & 65.00 & 62.50 \\
$\checkmark$ & $\checkmark$ & -- & -- & -- & -- & 50.24 & 79.89 & 19.90 & 24.68 & 51.79 & 58.70 & 26.19 & 48.33 & 44.97 \\
$\checkmark$ & $\checkmark$ & $\checkmark$ & -- & -- & -- & \underline{95.94} & 92.11 & \underline{33.34} & \underline{32.57} & 75.45 & 76.32 & \textbf{52.89} & 71.98 & \underline{66.33} \\
$\checkmark$ & $\checkmark$ & $\checkmark$ & $\checkmark$ & -- & -- & 56.80 & 64.01 & 26.50 & 27.36 & 56.40 & 60.80 & 43.50 & 63.10 & 49.81 \\
$\checkmark$ & $\checkmark$ & $\checkmark$ & $\checkmark$ & $\checkmark$ & -- & 95.34 & \underline{93.48} & 27.99 & 32.35 & \underline{77.28} & \underline{76.36} & 51.17 & \underline{72.61} & 65.82 \\
$\checkmark$ & $\checkmark$ & $\checkmark$ & $\checkmark$ & $\checkmark$ & $\checkmark$ & \textbf{96.51} & \textbf{94.45} & \textbf{34.85} & \textbf{33.51} & \textbf{78.07} & \textbf{77.14} & \underline{51.87} & \textbf{73.13} & \textbf{67.44} \\
\bottomrule
\end{tabular*}
\end{table*}

\section{Experiments}
\label{sec:experiments}

\subsection{Experimental Setup}
\label{sec:setup}

\noindent\textbf{Datasets.} We evaluate CoTeMIC on seven datasets, including Clevr4-10k~\cite{vaze2023no}, Fruit~\cite{hu2017finding}, Fruit360~\cite{yao2023augdmc}, CMUFace~\cite{gunnemann2014smvc}, Cards~\cite{yao2023augdmc}, CIFAR-10~\cite{krizhevsky2009learning}, and Flowers. Notably, Flowers is a new dataset we construct based on Flowers102~\cite{nilsback2008automated}. Detailed dataset statistics are in the supplementary material.


\noindent\textbf{Baselines.} We compare CoTeMIC with representative multiple image clustering methods, including five traditional methods (MSC~\cite{hu2017finding}, ENRC~\cite{miklautz2020deep}, iMClusts~\cite{ren2022diversified}, AugDMC~\cite{yao2023augdmc}, and DDMC~\cite{yao2024dual}), three MLLM-based methods (IC$\mid$TC~\cite{kwon2024image}, MLLMMC~\cite{zhang2026mllm} and ESMC~\cite{wang2026esmc}), and three VLM-based methods (Multi-MaP~\cite{yao2024multi}, Multi-Sub~\cite{yao2024customized}, and CRL~\cite{liu2025conditional}). These methods have been introduced in the Related Work section. For a fair comparison, we reproduce the MLLM-based and VLM-based methods using the same underlying MLLMs (LLaVA-1.5-7B~\cite{liu2024improved}) and VLMs (CLIP ViT-B/32~\cite{radford2021learning}), while keeping prompts and textual references as consistent as possible.

\noindent\textbf{Evaluation Metrics.} We primarily evaluate clustering performance using Normalized Mutual Information (NMI) and Rand Index (RI), which measure the agreement between predicted cluster assignments and ground-truth labels. For further analysis, we additionally employ Clustering Accuracy (ACC), which measures clustering correctness after optimal matching between predicted clusters and ground-truth classes. Higher values indicate better performance for all three metrics.

\noindent\textbf{Implementation Details.} We use GPT-6 to generate diverse attribute values and textual templates, from which we derive compositional text supervision for each clustering criterion. We adopt CLIP as the vision-language backbone, with both image and text encoders frozen throughout training. CoTeMIC is trained sequentially in three stages. First, we use the constructed Full-Target text pairs to train the text router. The text router is implemented as a two-layer MLP with a ReLU activation between the layers. Next, we freeze the text router and train the image adapter using two augmented views of each image. The image adapter employs a residual MLP, with the final linear layer in the residual branch initialized to zero. Finally, we freeze the text router and image adapter and train a linear clustering head.

For image augmentation, we apply random resized cropping with a scale range of $[0.8,1.0]$ and random horizontal flipping with a probability of $0.5$. All images are resized to $224\times224$ and normalized using CLIP's normalization statistics. All experiments are implemented in PyTorch on an NVIDIA RTX 5070 Ti GPU. We use the AdamW optimizer across all three stages, with a batch size of 128 and a weight decay of 0.01. All experiments are repeated with 10 random seeds, and the average results are reported.

\subsection{Performance Comparison}
\label{sec:comparison}
Table~\ref{tab:main_results} presents the clustering performance of CoTeMIC and competing methods across seven datasets and 16 clustering criteria. Overall, CoTeMIC achieves the best NMI and RI on most clustering criteria, demonstrating its robustness across diverse clustering scenarios.

On Clevr4-10k, CoTeMIC consistently outperforms the strongest baselines across all four criteria, achieving average improvements of 7.09\% and 2.20\% in NMI and RI, respectively. On Fruit360, its Species NMI reaches 69.31\%, surpassing the VLM-based method CRL by 31.88\%. Meanwhile, CoTeMIC achieves 78.04\% RI for Cards under the Suit criterion, outperforming the MLLM-based method ESMC by 5.19\%. Strong performance is also observed on Fruit, CMUFace, CIFAR-10, and Flowers, demonstrating the effectiveness of compositional text supervision for multiple image clustering.

\subsection{Ablation Study}
\label{sec:ablation}
\noindent\textbf{Loss Terms.} We investigate the contributions of six loss terms in CoTeMIC: text contrastive loss $\mathcal{L}_{\mathrm{TC}}$, image consistency loss $\mathcal{L}_{\mathrm{IC}}$, moment matching loss $\mathcal{L}_{\mathrm{MM}}$, confidence loss $\mathcal{L}_{\mathrm{Con}}$, balance loss $\mathcal{L}_{\mathrm{Bal}}$, and neighborhood consistency loss $\mathcal{L}_{\mathrm{NC}}$. For variants without clustering-related losses, we obtain cluster assignments using K-means. As shown in Table~\ref{tab:ablation}, $\mathcal{L}_{\mathrm{TC}}$ substantially improves performance over the CLIP-based K-means baseline, demonstrating the effectiveness of compositional text supervision. Adding $\mathcal{L}_{\mathrm{IC}}$ alone degrades performance, whereas $\mathcal{L}_{\mathrm{MM}}$ markedly improves it, highlighting the importance of moment matching for image adaptation. Similarly, $\mathcal{L}_{\mathrm{Con}}$ alone reduces performance, while $\mathcal{L}_{\mathrm{Bal}}$ mitigates this degradation, suggesting the importance of balancing cluster assignments. Finally, incorporating $\mathcal{L}_{\mathrm{NC}}$ achieves the best performance on seven of eight criteria, demonstrating the benefit of text-guided neighborhood consistency.

\noindent\textbf{Text Supervision.} To investigate the effectiveness of compositional text supervision in learning criterion-selective representation transformations, we compare three supervision strategies: Target-only, Full-only, and Full-Target (Ours). Target-only trains the text router using only target descriptions, where each positive pair consists of two descriptions with the same target attribute class but different templates. Full-only trains the text router using only full descriptions, where each positive pair consists of two descriptions that share the same target attribute class but differ in non-target attributes. In contrast, Full-Target constructs positive pairs by pairing full descriptions with their corresponding target descriptions. As shown in Table~\ref{tab:text_supervision}, Full-Target achieves the highest ACC across all six criteria, demonstrating the superiority of our proposed strategy.

\begin{table}[ht]
\centering
\caption{Ablation study of different text supervision strategies. Clustering accuracy (ACC, \%) is reported for each criterion. The best and second-best results are highlighted in \textbf{bold} and \underline{underline}, respectively.}
\label{tab:text_supervision}
\footnotesize
\setlength{\tabcolsep}{2.5pt}
\renewcommand{\arraystretch}{1}
\resizebox{\columnwidth}{!}{
\begin{tabular}{lcccccc}
\toprule
\multirow{2}{*}{Text Supervision} & \multicolumn{4}{c}{Clevr4-10k} & \multicolumn{2}{c}{Fruit360} \\
\cmidrule(lr){2-5}\cmidrule(lr){6-7}
& Shape & Color & Count & Texture & Species & Color \\
\midrule
Target-only & 94.18 & 93.41 & 31.98 & 26.04 & 68.06 & 60.32 \\
Full-only & \underline{94.27} & \underline{93.72} & \underline{32.95} & \underline{32.73} & \underline{76.39} & \underline{75.04} \\
Full-Target (Ours) & \textbf{96.51} & \textbf{94.45} & \textbf{34.85} & \textbf{33.51} & \textbf{78.07} & \textbf{77.14} \\
\bottomrule
\end{tabular}
}
\end{table}

\subsection{Criterion Selectivity Analysis}
\label{sec:criterion_selectivity}

Criterion-selective representation transformations should emphasize criterion-relevant semantic distinctions while suppressing criterion-irrelevant semantic variations. To quantitatively evaluate this property, we construct two types of image pairs for each criterion: pairs with different target attribute classes but identical non-target attribute classes, and pairs with identical target attribute classes but different non-target attribute classes. For each method, we use the same image pairs and compute the mean cosine distance in the representation space for each pair type. The resulting distances, denoted by $D_{\mathrm{target}}$ and $D_{\mathrm{non}}$, reflect criterion-relevant semantic distinctions and criterion-irrelevant semantic variations, respectively. Since the magnitude of cosine distances may vary across representation spaces, we report the ratio $r = D_{\mathrm{target}} / D_{\mathrm{non}}$, with a higher ratio indicating greater criterion selectivity. As shown in Table~\ref{tab:criterion-selectivity}, CoTeMIC achieves the highest ratio across all criteria, demonstrating superior criterion selectivity.

\begin{table}[ht]
\centering
\caption{Comparison of criterion selectivity measured by the distance ratio $r$. The best and second-best results are highlighted in \textbf{bold} and \underline{underline}, respectively.}
\label{tab:criterion-selectivity}
\footnotesize
\setlength{\tabcolsep}{4.5pt}
\renewcommand{\arraystretch}{1}
\resizebox{\columnwidth}{!}{
\begin{tabular}{lcccccc}
\toprule
\multirow{2}{*}{Method} & \multicolumn{2}{c}{Cards} & \multicolumn{2}{c}{Clevr4-10k} & \multicolumn{2}{c}{CMUFace} \\
\cmidrule(lr){2-3}\cmidrule(lr){4-5}\cmidrule(lr){6-7}
& Suit & Number & Color & Shape & Glass & Pose \\
\midrule
CLIP & 0.99 & 1.02 & 0.80 & 1.25 & 1.04 & 0.96 \\
CRL & \underline{1.70} & \underline{1.67} & \underline{3.60} & \underline{2.58} & \underline{4.19} & \underline{1.84} \\
Multi-Sub & 1.27 & 1.49 & 1.46 & 1.73 & 1.13 & 0.90 \\
Multi-MaP & 1.43 & 1.45 & 2.95 & 1.81 & 1.24 & 1.19 \\
CoTeMIC & \textbf{1.96} & \textbf{1.86} & \textbf{6.26} & \textbf{4.35} & \textbf{23.60} & \textbf{2.08} \\
\bottomrule
\end{tabular}
}
\end{table}

\subsection{Representation Visualization}
\label{sec:representation}
\begin{figure}[ht]
    \centering
    \includegraphics[width=\linewidth]{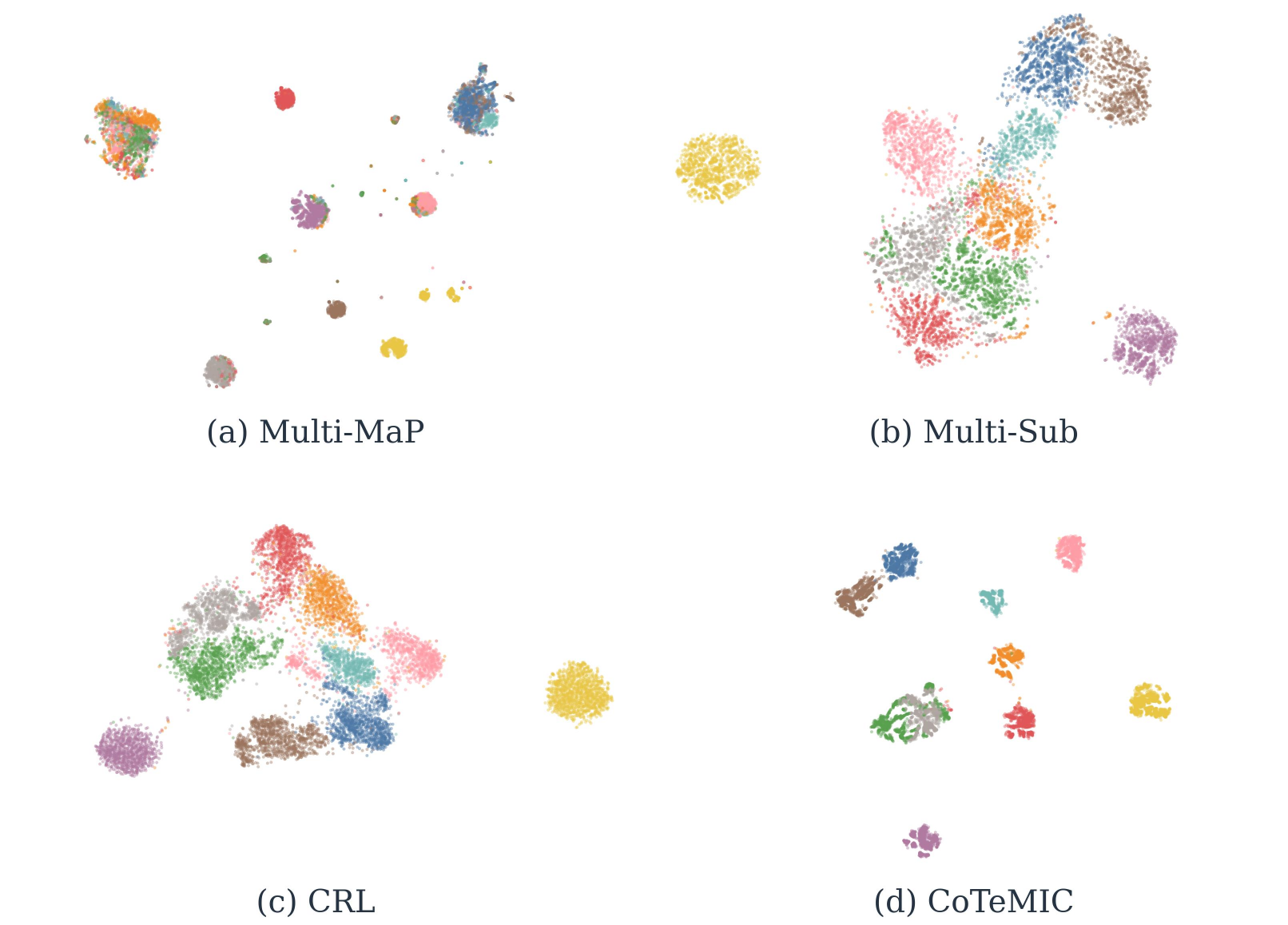}
    \caption{t-SNE visualization of criterion-specific representations on Clevr4-10k under the Shape criterion. Colors denote shape categories. Compared with the baselines (a, b, c), CoTeMIC (d) exhibits more distinct cluster structures.}
    \label{fig:tsne}
\end{figure}

To qualitatively evaluate the learned representations, we use t-SNE to visualize the criterion-specific representations obtained by different methods on Clevr4-10k under the Shape criterion. Colors denote different ground-truth shape categories. As shown in Fig.~\ref{fig:tsne}, Multi-MaP (a) exhibits mixing of different classes within the same clusters, whereas Multi-Sub (b) and CRL (c) show poorly defined boundaries between neighboring clusters. In contrast, CoTeMIC (d) exhibits better intra-class compactness and inter-class separation, yielding a more discriminative representation space under the specified criterion.

\subsection{Generalization Across VLM Backbones}
\label{sec:backbone_generalization}

We evaluate CoTeMIC across three pretrained VLM backbones (CLIP, ALIGN, and SigLIP) on Cards and Clevr4-10k. We compare CoTeMIC with K-means and CRL. As shown in Table~\ref{tab:vlm_backbones}, CoTeMIC outperforms both baselines across all three backbones, improving mean ACC over CRL by 10.86\%, 8.75\%, and 8.94\%, respectively. These consistent improvements demonstrate the generalization ability of CoTeMIC across different pretrained VLM backbones.

\begin{table}[ht]
\centering
\caption{Comparison of clustering performance across VLM backbones. Clustering accuracy (ACC, \%) is reported for each criterion. The best and second-best available results are highlighted in \textbf{bold} and \underline{underline}, respectively.}
\label{tab:vlm_backbones}
\begin{footnotesize}
\setlength{\tabcolsep}{3.5pt}
\renewcommand{\arraystretch}{1}
\resizebox{\columnwidth}{!}{
\begin{tabular}{lccccc}
\toprule
\multirow{2}{*}{Method} & \multicolumn{2}{c}{Clevr4-10k} & \multicolumn{2}{c}{Cards} & \multirow{2}{*}{Mean} \\
\cmidrule(lr){2-3}\cmidrule(lr){4-5}
& Shape & Color & Number & Suit & \\
\midrule
CLIP+K-means & 77.95 & 11.96 & 18.50 & 44.55 & 38.24 \\
CLIP+CRL & \underline{81.48} & \underline{90.68} & \underline{34.81} & \underline{65.55} & \underline{68.13} \\
CLIP+CoTeMIC & \textbf{96.51} & \textbf{94.45} & \textbf{51.87} & \textbf{73.13} & \textbf{78.99} \\
\midrule
ALIGN+K-means & \underline{94.23} & 11.56 & 18.15 & 31.92 & 38.97 \\
ALIGN+CRL & 89.02 & \underline{83.95} & \underline{29.34} & \underline{61.00} & \underline{65.83} \\
ALIGN+CoTeMIC & \textbf{94.29} & \textbf{86.33} & \textbf{39.80} & \textbf{77.89} & \textbf{74.58} \\
\midrule
SigLIP+K-means & \underline{93.97} & 11.43 & 35.18 & 43.47 & 46.01 \\
SigLIP+CRL & 92.97 & \underline{82.27} & \underline{82.86} & \underline{75.37} & \underline{83.37} \\
SigLIP+CoTeMIC & \textbf{95.48} & \textbf{89.87} & \textbf{89.11} & \textbf{94.76} & \textbf{92.31} \\
\bottomrule
\end{tabular}
}
\end{footnotesize}
\end{table}

\section{Conclusion}
In this paper, we proposed CoTeMIC, a compositional text-supervised framework for multiple image clustering. By leveraging Full–Target text pairs as compositional text supervision, CoTeMIC learns criterion-selective transformations. Furthermore, a consistency-guided image adapter facilitates the application of these transformations to the visual domain, while a clustering head produces criterion-specific partitions. Extensive experiments across diverse datasets demonstrate the effectiveness and generalization ability of CoTeMIC in multiple image clustering.


\small
\bibliographystyle{ieeenat_fullname}
\bibliography{main}

\clearpage
\setcounter{page}{1}
\maketitlesupplementary

\section{Dataset Details}
\label{sec:supp-datasets}

We evaluate CoTeMIC on 7 datasets, each with multiple clustering criteria that group the same images by different attributes.

\subsection{Clevr4-10k}
\label{sec:supp-clevr}
\begin{figure}[h]
    \centering
    \includegraphics[width=\linewidth]{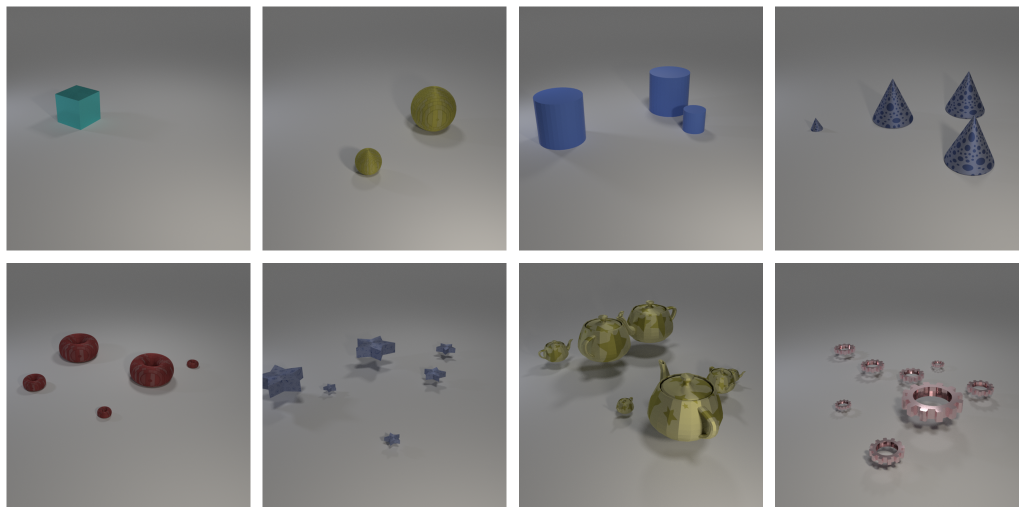}
    \caption{Sample images from Clevr4-10k.}
    \label{fig:supp-examples-clevr4-10k}
\end{figure}

The Clevr4-10k dataset~\citep{vaze2023no} contains 10,531 synthetic images of scenes populated with 3D objects.
Each scene is labeled by object shape, color, count, and surface texture.
Under the Count criterion, scenes with the same number of objects belong to the same class even when those objects differ in shape or color.
Fig.~\ref{fig:supp-examples-clevr4-10k} shows sample images, and Table~\ref{tab:supp-clevr4-10k} lists the 10 classes for each criterion.

\begin{table}[h]
\centering
\caption{Structure of the Clevr4-10k dataset.}
\label{tab:supp-clevr4-10k}
\small
\setlength{\tabcolsep}{3pt}
\begin{tabular*}{\linewidth}{@{}l@{\extracolsep{\fill}}cp{0.50\linewidth}@{}}
\toprule
Criterion & \shortstack{Number of Classes} & Class Examples \\
\midrule
Shape & 10 & cube, sphere, monkey, cone, torus, star, teapot, diamond, gear, cylinder \\
Color & 10 & gray, red, blue, green, brown, purple, cyan, yellow, pink, orange \\
Count & 10 & one, two, three, four, five, six, seven, eight, nine, ten \\
Texture & 10 & rubber, metal, checkered, emojis, wave, brick, star, circles, zigzag, chessboard \\
\bottomrule
\end{tabular*}
\end{table}

\subsection{Fruit}
\label{sec:supp-fruit}
\begin{figure}[h]
    \centering
    \includegraphics[width=\linewidth]{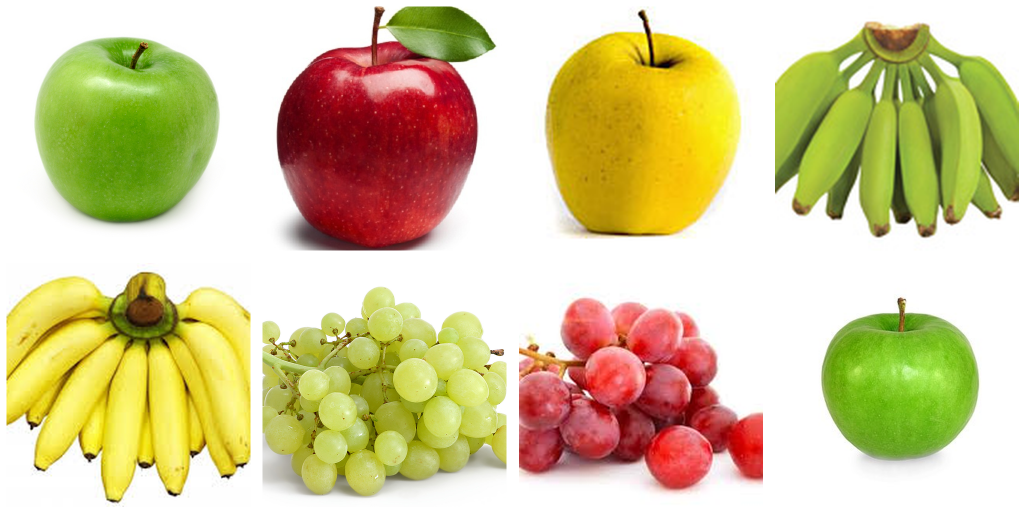}
    \caption{Sample images from Fruit.}
    \label{fig:supp-examples-fruit}
\end{figure}

The Fruit dataset~\citep{hu2017finding} contains 105 photographs of apples, bananas, and grapes.
Fruits of the same species can differ in color, and fruits of different species can share a color label.
These variations support grouping the same images by either species or color.
Fig.~\ref{fig:supp-examples-fruit} shows sample images, and Table~\ref{tab:supp-fruit} lists the 3 classes for each criterion.

\begin{table}[h]
\centering
\caption{Structure of the Fruit dataset.}
\label{tab:supp-fruit}
\small
\setlength{\tabcolsep}{3pt}
\begin{tabular*}{\linewidth}{@{}l@{\extracolsep{\fill}}cp{0.50\linewidth}@{}}
\toprule
Criterion & \shortstack{Number of Classes} & Class Examples \\
\midrule
Species & 3 & apple, banana, grape \\
Color & 3 & green, red, yellow \\
\bottomrule
\end{tabular*}
\end{table}

\subsection{Fruit360}
\label{sec:supp-fruit360}
\begin{figure}[h]
    \centering
    \includegraphics[width=\linewidth]{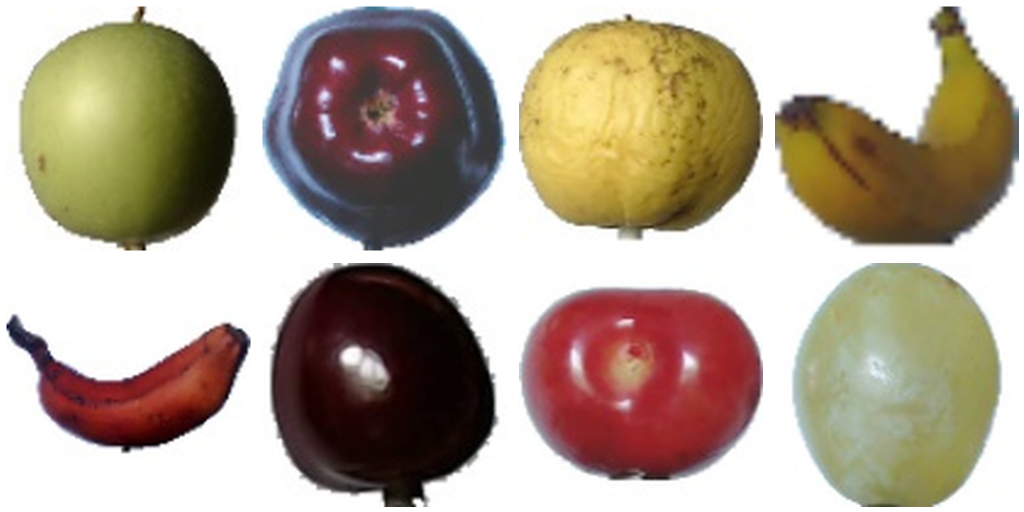}
    \caption{Sample images from Fruit360.}
    \label{fig:supp-examples-fruit360}
\end{figure}

The Fruit360 dataset~\citep{yao2023augdmc} used in our experiments contains 4,856 fruit photographs.
The Species criterion covers apples, bananas, cherries, and grapes.
The Color criterion distinguishes 4 colors, including the visually similar burgundy and red.
Fig.~\ref{fig:supp-examples-fruit360} shows sample images, and Table~\ref{tab:supp-fruit360} lists the classes for both criteria.

\begin{table}[h]
\centering
\caption{Structure of the Fruit360 dataset.}
\label{tab:supp-fruit360}
\small
\setlength{\tabcolsep}{3pt}
\begin{tabular*}{\linewidth}{@{}l@{\extracolsep{\fill}}cp{0.50\linewidth}@{}}
\toprule
Criterion & \shortstack{Number of Classes} & Class Examples \\
\midrule
Species & 4 & apple, banana, cherry, grape \\
Color & 4 & burgundy, green, red, yellow \\
\bottomrule
\end{tabular*}
\end{table}

\subsection{CMUFace}
\label{sec:supp-cmuface}
\begin{figure}[h]
    \centering
    \includegraphics[width=\linewidth]{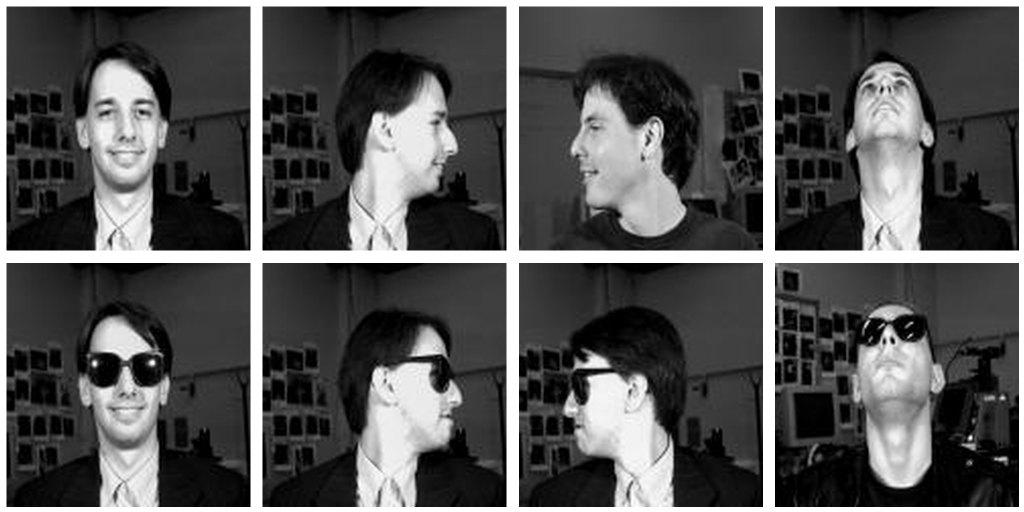}
    \caption{Sample images from CMUFace.}
    \label{fig:supp-examples-cmuface}
\end{figure}

The CMUFace dataset~\citep{gunnemann2014smvc} contains 624 face photographs that vary in head orientation and eyewear.
The Pose criterion distinguishes 4 head orientations, while the Glass criterion groups images according to whether the subject wears sunglasses.
Fig.~\ref{fig:supp-examples-cmuface} shows sample images, and Table~\ref{tab:supp-cmuface} lists the classes for both criteria.

\vspace{-1em}

\begin{table}[h]
\centering
\caption{Structure of the CMUFace dataset.}
\label{tab:supp-cmuface}
\small
\setlength{\tabcolsep}{3pt}
\begin{tabular*}{\linewidth}{@{}l@{\extracolsep{\fill}}cp{0.50\linewidth}@{}}
\toprule
Criterion & \shortstack{Number of Classes} & Class Examples \\
\midrule
Pose & 4 & left, right, straight, up \\
Glass & 2 & open, sunglasses \\
\bottomrule
\end{tabular*}
\end{table}

\subsection{Cards}
\label{sec:supp-cards}
\begin{figure}[h]
    \centering
    \includegraphics[width=\linewidth]{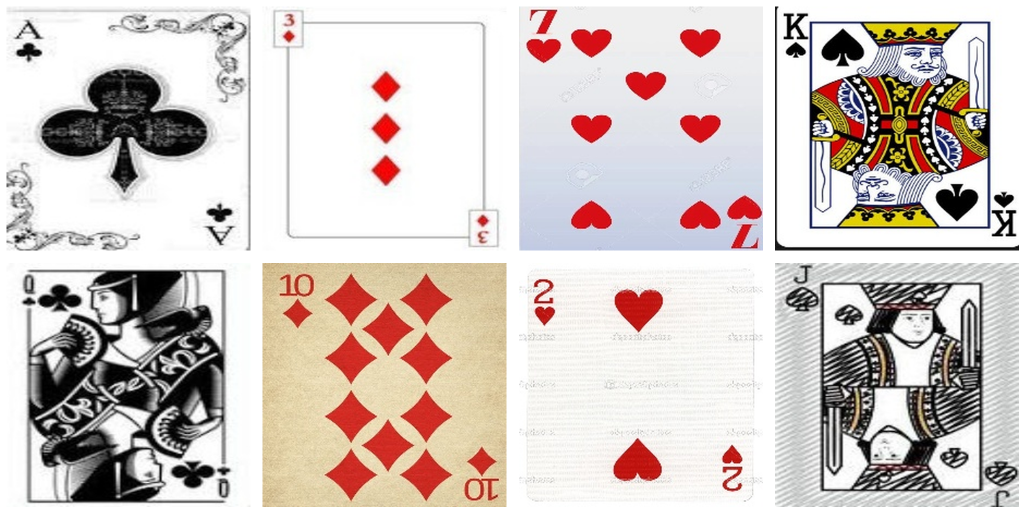}
    \caption{Sample images from Cards.}
    \label{fig:supp-examples-cards}
\end{figure}

The Cards dataset~\citep{yao2023augdmc} contains 8,029 photographs of playing cards taken under different lighting conditions, from different angles, and against different backgrounds.
The Number criterion groups cards of the same rank across suits, while the Suit criterion groups cards of the same suit across ranks.
Fig.~\ref{fig:supp-examples-cards} shows sample images, and Table~\ref{tab:supp-cards} lists the 13 ranks and 4 suits.

\begin{table}[h]
\centering
\caption{Structure of the Cards dataset.}
\label{tab:supp-cards}
\small
\setlength{\tabcolsep}{3pt}
\begin{tabular*}{\linewidth}{@{}l@{\extracolsep{\fill}}cp{0.50\linewidth}@{}}
\toprule
Criterion & \shortstack{Number of Classes} & Class Examples \\
\midrule
Number & 13 & ace, two, three, four, five, six, seven, eight, nine, ten, jack, queen, king \\
Suit & 4 & clubs~($\clubsuit$), diamonds~($\diamondsuit$), hearts~($\heartsuit$), spades~($\spadesuit$) \\
\bottomrule
\end{tabular*}
\end{table}

\subsection{CIFAR-10}
\label{sec:supp-cifar10}
\begin{figure}[h]
    \centering
    \includegraphics[width=\linewidth]{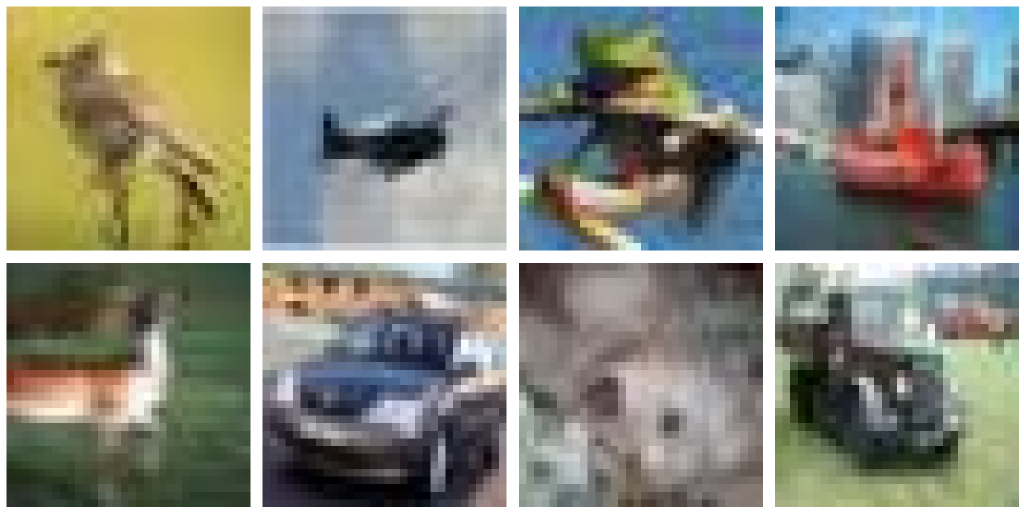}
    \caption{Sample images from CIFAR-10.}
    \label{fig:supp-examples-cifar10}
\end{figure}

The CIFAR-10 dataset~\citep{krizhevsky2009learning} provides 50,000 color images for our evaluation.
The Type criterion distinguishes animals from vehicles, while the Environment criterion groups images according to their associated environments.
For example, cars and airplanes share the vehicle label under Type but belong to the land and air classes, respectively, under Environment.
Fig.~\ref{fig:supp-examples-cifar10} shows sample images, and Table~\ref{tab:supp-cifar-10} lists the classes for both criteria.

\begin{table}[h]
    \centering
    \caption{Structure of the CIFAR-10 dataset.}
    \label{tab:supp-cifar-10}
    \small
    \setlength{\tabcolsep}{3pt}
    \begin{tabular*}{\linewidth}{@{}l@{\extracolsep{\fill}}cp{0.50\linewidth}@{}}
        \toprule
        Criterion & \shortstack{Number of Classes} & Class Examples \\
        \midrule
        Type & 2 & animal, vehicle \\
        Environment & 3 & land, air, water \\
        \bottomrule
    \end{tabular*}
\end{table}

\subsection{Flowers}
\label{sec:supp-flowers}
\begin{figure}[h]
    \centering
    \includegraphics[width=\linewidth]{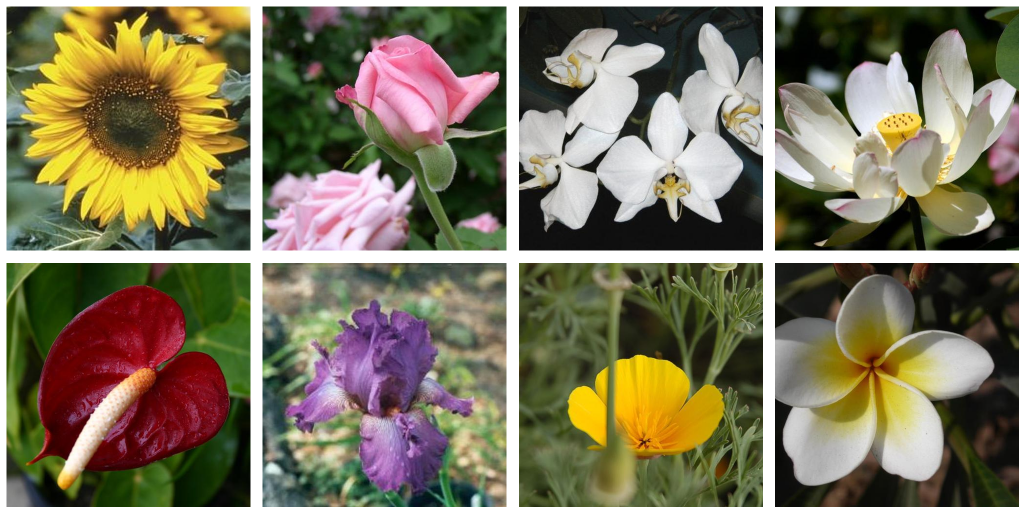}
    \caption{Sample images from Flowers.}
    \label{fig:supp-examples-flowers}
\end{figure}

The Flowers dataset, which we constructed from Flowers102~\citep{nilsback2008automated}, contains 1,756 images spanning 40 species and 5 blossom colors.
Under the Species criterion, yellow and white primroses belong to the same class.
Under the Color criterion, a yellow primrose and a yellow dandelion belong to the same class.
Fig.~\ref{fig:supp-examples-flowers} shows sample images, and Table~\ref{tab:supp-flowers} lists the species and color classes.

\begin{table}[h]
\centering
\caption{Structure of the Flowers dataset.}
\label{tab:supp-flowers}
\small
\setlength{\tabcolsep}{3pt}
\begin{tabular*}{\linewidth}{@{}l@{\extracolsep{\fill}}cp{0.50\linewidth}@{}}
\toprule
Criterion & \shortstack{Number of\\Classes} & Class Examples \\
\midrule
Species & 40 & anthurium, artichoke, azalea, bearded iris, bee balm, dahlia, buttercup, California poppy, Canterbury bells, cape flower, carnation, clematis, dandelion, cyclamen, daffodil, foxglove, frangipani, geranium, white arum lily, globeflower, hibiscus, lotus, magnolia, mallow, marigold, moon orchid, oxeye daisy, petunia, pincushion flower, poinsettia, primrose, red ginger, rose, snapdragon, spring crocus, sunflower, gladiolus, thorn apple, wallflower, watercress \\
Color & 5 & pink, purple, red, white, yellow \\
\bottomrule
\end{tabular*}
\end{table}

\section{Compositional Text Supervision}
\label{sec:supp-compositional-text}

We illustrate the construction of Full-Target text pairs described in Sec.~\ref{sec:router} using Fruit (Color), Flowers (Species), and Cards (Suit). For each dataset, we employ an LLM to generate attribute values for each criterion and enumerate all possible compositions across criteria, with each composition selecting one value per criterion. We then instantiate the corresponding templates to construct a full description containing all selected attribute values and a target description containing only the attribute value of the specified criterion. During text router training, we treat the full and target descriptions within each pair as positives and select negatives only from descriptions with different target attribute values, excluding those that share the anchor's target value to avoid false negatives. Notably, the number of attribute values used for text router training need not match the number of ground-truth classes used for image clustering. For example, on Fruit, we use 20 color values to train the text router, whereas image clustering is evaluated against 3 ground-truth color classes.

\subsection{Fruit (Color)}
\label{sec:supp-text-fruit-color}

For color-based clustering on Fruit, we use 20 color terms as target values
and 20 fruit species names as non-target values. Applying all 10 template
pairs to each color-species composition yields 4,000 Full-Target text pairs.

\subsubsection{Target Attribute (Color)}
\begingroup
\small
\raggedright

\noindent\textbf{LLM prompt.}\enspace
\texttt{Please generate common lowercase English expressions to describe
the skin color of fruit, as many as possible, formatted as:
["...", "...", "..."]. Ensure all items are unique and written in a
single line, without any nested lists or additional formatting. Only
generate the list, and do not include any additional information.}\par

\smallskip
\noindent\textbf{LLM response.}\enspace
\texttt{["burgundy", "green", "red", "yellow", "blue", "orange",
"purple", "pink", "brown", "black", "white", "gray", "cyan", "magenta",
"teal", "navy", "olive", "beige", "coral", "gold"]}\par
\endgroup

\subsubsection{Non-target Attribute (Species)}
\begingroup
\small
\raggedright

\noindent\textbf{LLM prompt.}\enspace
\texttt{Please generate common lowercase English names for fruits, as
many as possible, formatted as: ["...", "...", "..."]. Ensure all items
are unique and written in a single line, without any nested lists or
additional formatting. Only generate the list, and do not include any
additional information.}\par

\smallskip
\noindent\textbf{LLM response.}\enspace
\texttt{["apple", "banana", "cherry", "grape", "orange", "pear",
"peach", "plum", "lemon", "lime", "strawberry", "raspberry", "blueberry",
"blackberry", "pineapple", "mango", "kiwi", "watermelon", "cantaloupe",
"papaya"]}\par
\endgroup

\subsubsection{Full-Target Templates}
\begingroup
\small
\noindent
\begin{tabular}{@{}p{0.57\linewidth}@{\hspace{0.03\linewidth}}p{0.40\linewidth}@{}}
\toprule
\textbf{Full template} & \textbf{Target template}\tabularnewline
\midrule
\raggedright A photo shows a \verb|{species}|. The fruit is \verb|{color}|.
    & \raggedright The fruit is \verb|{color}|.\tabularnewline[2pt]
\raggedright The image contains a \verb|{species}| with \verb|{color}| skin.
    & \raggedright The color of the fruit is \verb|{color}|.\tabularnewline[2pt]
\raggedright This is a photograph of a \verb|{species}|. Its surface is \verb|{color}|.
    & \raggedright The fruit has \verb|{color}| skin.\tabularnewline[2pt]
\raggedright Here is a \verb|{species}|, and its skin is \verb|{color}|.
    & \raggedright Its surface color is \verb|{color}|.\tabularnewline[2pt]
\raggedright The camera captures a \verb|{species}|; its color is \verb|{color}|.
    & \raggedright The visible fruit color is \verb|{color}|.\tabularnewline[2pt]
\raggedright The picture features a \verb|{species}| displaying \verb|{color}| skin.
    & \raggedright The fruit appears \verb|{color}|.\tabularnewline[2pt]
\raggedright A studio photograph presents a \verb|{species}| with \verb|{color}| coloration.
    & \raggedright The fruit's exterior is \verb|{color}|.\tabularnewline[2pt]
\raggedright In this image, the fruit is a \verb|{species}|. Its peel looks \verb|{color}|.
    & \raggedright The skin of the fruit looks \verb|{color}|.\tabularnewline[2pt]
\raggedright The subject of the photograph is a \verb|{species}|. Its skin looks \verb|{color}|.
    & \raggedright The depicted color is \verb|{color}|.\tabularnewline[2pt]
\raggedright A detailed photograph shows a \verb|{species}|. The visible fruit color is \verb|{color}|.
    & \raggedright The fruit shows \verb|{color}| coloration.\tabularnewline
\bottomrule
\end{tabular}\par
\endgroup

\subsubsection{Example Full-Target Pair}
\noindent\textbf{Full description.}\enspace
A photo shows a cherry. The fruit is red.\par
\noindent\textbf{Target description.}\enspace
The fruit is red.\par

\subsection{Flowers (Species)}
\label{sec:supp-text-flowers-species}

For species-based clustering on Flowers, we use 80 flower names as target values
and 20 blossom color terms as non-target values. Applying all 10 template pairs
to each species-color composition yields 16,000 Full-Target text pairs.

\subsubsection{Target Attribute (Species)}
\begingroup
\small
\raggedright

\noindent\textbf{LLM prompt.}\enspace
\texttt{Please generate common lowercase English names for flowering plants, as many
as possible, formatted as: ["...", "...", "..."]. Ensure all items are unique
and written in a single line, without any nested lists or additional
formatting. Only generate the list, and do not include any additional
information.}\par

\smallskip
\noindent\textbf{LLM response.}\enspace
\texttt{["anthurium", "artichoke", "azalea", "bearded iris", "bee balm", "dahlia",
"buttercup", "california poppy", "canterbury bells", "nerine", "carnation",
"clematis", "dandelion", "cyclamen", "daffodil", "foxglove", "frangipani",
"geranium", "calla lily", "globeflower", "hibiscus", "lotus", "magnolia",
"flowering maple", "marigold", "moth orchid", "oxeye daisy", "petunia",
"pincushion flower", "poinsettia", "primrose", "red ginger", "rose",
"snapdragon", "crocus", "sunflower", "gladiolus", "thorn apple",
"wallflower", "nasturtium", "slipper orchid", "sweet pea", "tiger lily",
"bird of paradise", "monkshood", "globe thistle", "king protea",
"coneflower", "peruvian lily", "balloon flower", "glory lily", "fritillary",
"grape hyacinth", "corn poppy", "garden phlox", "love-in-a-mist", "cosmos",
"sea holly", "cattleya orchid", "siam tulip", "hellebore", "gerbera daisy",
"lisianthus", "pansy", "gaura", "japanese anemone", "black-eyed susan",
"gazania", "water lily", "passion flower", "toad lily", "columbine", "desert
rose", "canna lily", "amaryllis", "bougainvillea", "bromeliad", "blanket
flower", "trumpet vine", "blackberry lily"]}\par
\endgroup

\subsubsection{Non-target Attribute (Color)}
\begingroup
\small
\raggedright

\noindent\textbf{LLM prompt.}\enspace
\texttt{Please generate common lowercase English expressions to describe blossom
colors, as many as possible, formatted as: ["...", "...", "..."]. Ensure all
items are unique and written in a single line, without any nested lists or
additional formatting. Only generate the list, and do not include any
additional information.}\par

\smallskip
\noindent\textbf{LLM response.}\enspace
\texttt{["burgundy", "green", "red", "yellow", "blue", "orange", "purple", "pink",
"brown", "black", "white", "gray", "cyan", "magenta", "teal", "navy",
"olive", "beige", "coral", "gold"]}\par
\endgroup

\subsubsection{Full-Target Templates}
\begingroup
\small
\noindent
\begin{tabular}{@{}p{0.57\linewidth}@{\hspace{0.03\linewidth}}p{0.40\linewidth}@{}}
\toprule
\textbf{Full template} & \textbf{Target template}\tabularnewline
\midrule
\raggedright A photograph of \verb|{species}| with \verb|{color}| flowers.
    & \raggedright The plant is \verb|{species}|.\tabularnewline[2pt]
\raggedright The image shows \verb|{species}| in bloom. Its blossoms are \verb|{color}|.
    & \raggedright The image shows \verb|{species}|.\tabularnewline[2pt]
\raggedright A close-up shows \verb|{species}| with \verb|{color}| petals.
    & \raggedright A close-up of \verb|{species}|.\tabularnewline[2pt]
\raggedright The photograph features \verb|{species}| bearing \verb|{color}| blooms.
    & \raggedright The photograph features \verb|{species}|.\tabularnewline[2pt]
\raggedright Here we see \verb|{species}|. The flowers have \verb|{color}| coloration.
    & \raggedright Here we see \verb|{species}|.\tabularnewline[2pt]
\raggedright The camera captures \verb|{species}| displaying \verb|{color}| blossoms.
    & \raggedright The camera captures \verb|{species}|.\tabularnewline[2pt]
\raggedright This picture presents \verb|{species}| with flowers that are \verb|{color}|.
    & \raggedright This picture presents \verb|{species}|.\tabularnewline[2pt]
\raggedright The flowering plant is \verb|{species}|, and its blossoms look \verb|{color}|.
    & \raggedright The flowering plant is \verb|{species}|.\tabularnewline[2pt]
\raggedright A detailed photograph shows \verb|{species}|. Its flower color is \verb|{color}|.
    & \raggedright A detailed photograph of \verb|{species}|.\tabularnewline[2pt]
\raggedright The pictured \verb|{species}| has \verb|{color}| blooms.
    & \raggedright The pictured plant is \verb|{species}|.\tabularnewline
\bottomrule
\end{tabular}\par
\endgroup

\subsubsection{Example Full-Target Pair}
\noindent\textbf{Full description.}\enspace
A photograph of hibiscus with red flowers.\par
\noindent\textbf{Target description.}\enspace
The plant is hibiscus.\par

\subsection{Cards (Suit)}
\label{sec:supp-text-cards-suit}

For suit-based clustering on Cards, we use 4 suit names as target values and
13 card ranks as non-target values. Applying all 50 template pairs to each
rank-suit composition yields 2,600 Full-Target text pairs.

\subsubsection{Target Attribute (Suit)}
\begingroup
\small
\raggedright

\noindent\textbf{LLM prompt.}\enspace
\texttt{Please list all 4 suits in a standard 52-card deck using their conventional
lowercase English names, formatted as: ["...", "...", "..."]. Ensure all
items are unique and written in a single line, without any nested lists or
additional formatting. Only generate the list, and do not include any
additional information.}\par

\smallskip
\noindent\textbf{LLM response.}\enspace
\texttt{["clubs", "diamonds", "hearts", "spades"]}\par
\endgroup

\subsubsection{Non-target Attribute (Number)}
\begingroup
\small
\raggedright

\noindent\textbf{LLM prompt.}\enspace
\texttt{Please list all 13 ranks in a standard 52-card deck using their conventional
lowercase English names, formatted as: ["...", "...", "..."]. Ensure all
items are unique and written in a single line, without any nested lists or
additional formatting. Only generate the list, and do not include any
additional information.}\par

\smallskip
\noindent\textbf{LLM response.}\enspace
\texttt{["ace", "two", "three", "four", "five", "six", "seven", "eight", "nine",
"ten", "jack", "queen", "king"]}\par
\endgroup

\begin{table*}[t]
\centering
\caption{CoTeMIC clustering performance across 10 random seeds. Means and standard deviations are reported as percentages.}
\label{tab:supp-main-seed-results}
\footnotesize
\setlength{\tabcolsep}{1.2pt}
\begin{tabular*}{\textwidth}{@{\extracolsep{\fill}}cc*{16}{c}@{}}
\toprule
\multirow{2}{*}{Metric} & \multirow{2}{*}{Statistic} & \multicolumn{4}{c}{Clevr4-10k} & \multicolumn{2}{c}{Fruit} & \multicolumn{2}{c}{Fruit360} & \multicolumn{2}{c}{CMUFace} & \multicolumn{2}{c}{Cards} & \multicolumn{2}{c}{CIFAR-10} & \multicolumn{2}{c}{Flowers} \\
\cmidrule(lr){3-6} \cmidrule(lr){7-8} \cmidrule(lr){9-10} \cmidrule(lr){11-12} \cmidrule(lr){13-14} \cmidrule(lr){15-16} \cmidrule(lr){17-18}
 & & Shape & Color & Count & Texture & Species & Color & Species & Color & Pose & Glass & Number & Suit & Type & Env. & Species & Color \\
\midrule
\multirow{2}{*}{NMI} & Mean & 94.98 & 92.44 & 30.20 & 22.10 & 100.00 & 90.64 & 69.31 & 66.39 & 51.09 & 93.08 & 44.28 & 40.90 & 80.54 & 40.03 & 91.22 & 95.16 \\
 & SD & 1.90 & 0.80 & 0.75 & 1.02 & 0.00 & 11.95 & 6.43 & 6.07 & 1.65 & 2.04 & 1.05 & 2.67 & 0.36 & 3.08 & 0.75 & 1.52 \\
\midrule
\multirow{2}{*}{ACC} & Mean & 96.51 & 94.45 & 34.85 & 33.51 & 100.00 & 96.29 & 78.07 & 77.14 & 68.64 & 99.12 & 51.87 & 73.13 & 97.07 & 79.34 & 84.09 & 98.08 \\
 & SD & 1.72 & 0.63 & 0.72 & 0.89 & 0.00 & 7.16 & 4.64 & 6.55 & 1.34 & 0.34 & 1.92 & 3.38 & 0.07 & 2.51 & 2.65 & 1.91 \\
\midrule
\multirow{2}{*}{RI} & Mean & 98.74 & 98.13 & 85.60 & 84.74 & 100.00 & 95.90 & 83.39 & 83.63 & 80.47 & 98.25 & 90.80 & 78.04 & 94.32 & 74.83 & 98.85 & 98.70 \\
 & SD & 0.56 & 0.22 & 0.09 & 0.14 & 0.00 & 6.65 & 3.20 & 2.91 & 0.63 & 0.67 & 0.21 & 1.49 & 0.13 & 2.54 & 0.15 & 0.60 \\
\bottomrule
\end{tabular*}
\end{table*}

\subsubsection{Full-Target Templates}
\begingroup
\small
\setlength{\parskip}{2pt}
\newcommand{\CardsTemplateSlot}[1]{\texttt{\detokenize{{#1}}}}
\newcommand{\CardsTemplateRow}[2]{%
    \nointerlineskip
    \noindent
    \begin{tabular}[t]{@{}p{0.57\linewidth}@{\hspace{0.03\linewidth}}p{0.40\linewidth}@{}}
        \raggedright #1 & \raggedright #2\tabularnewline
    \end{tabular}\par
}
\hrule height \heavyrulewidth
\nobreak
\CardsTemplateRow{\textbf{Full template}}{\textbf{Target template}}
\nobreak
\hrule height \lightrulewidth
\nobreak
\CardsTemplateRow{The image presents the \CardsTemplateSlot{number} of \CardsTemplateSlot{suit}.}{Its suit is \CardsTemplateSlot{suit}.}
\CardsTemplateRow{At the center lies the \CardsTemplateSlot{number} of \CardsTemplateSlot{suit}.}{The card belongs to \CardsTemplateSlot{suit}.}
\CardsTemplateRow{The photograph clearly shows the \CardsTemplateSlot{number} of \CardsTemplateSlot{suit}.}{The visible suit is \CardsTemplateSlot{suit}.}
\CardsTemplateRow{A detailed view reveals the \CardsTemplateSlot{number} of \CardsTemplateSlot{suit}.}{The suit is \CardsTemplateSlot{suit}.}
\CardsTemplateRow{The pictured playing card is the \CardsTemplateSlot{number} of \CardsTemplateSlot{suit}.}{The pictured suit is \CardsTemplateSlot{suit}.}
\CardsTemplateRow{Clearly visible is the \CardsTemplateSlot{number} of \CardsTemplateSlot{suit}.}{Its suit is \CardsTemplateSlot{suit}.}
\CardsTemplateRow{The scene contains the \CardsTemplateSlot{number} of \CardsTemplateSlot{suit}.}{The card uses the \CardsTemplateSlot{suit} suit.}
\CardsTemplateRow{This close-up captures the \CardsTemplateSlot{number} of \CardsTemplateSlot{suit}.}{The close-up emphasizes \CardsTemplateSlot{suit}.}
\CardsTemplateRow{The card on display is the \CardsTemplateSlot{number} of \CardsTemplateSlot{suit}.}{Its displayed suit is \CardsTemplateSlot{suit}.}
\CardsTemplateRow{In the image, the \CardsTemplateSlot{number} of \CardsTemplateSlot{suit} appears.}{The card belongs to \CardsTemplateSlot{suit}.}
\CardsTemplateRow{A single \CardsTemplateSlot{number} of \CardsTemplateSlot{suit} is visible.}{It bears the \CardsTemplateSlot{suit} suit.}
\CardsTemplateRow{The frame features the \CardsTemplateSlot{number} of \CardsTemplateSlot{suit}.}{The featured suit is \CardsTemplateSlot{suit}.}
\CardsTemplateRow{The visible playing card is the \CardsTemplateSlot{number} of \CardsTemplateSlot{suit}.}{The visible suit is \CardsTemplateSlot{suit}.}
\CardsTemplateRow{Shown against the background is the \CardsTemplateSlot{number} of \CardsTemplateSlot{suit}.}{Its suit can be identified as \CardsTemplateSlot{suit}.}
\CardsTemplateRow{The composition highlights the \CardsTemplateSlot{number} of \CardsTemplateSlot{suit}.}{The highlighted suit is \CardsTemplateSlot{suit}.}
\CardsTemplateRow{The card face displays the \CardsTemplateSlot{number} of \CardsTemplateSlot{suit}.}{The face displays the \CardsTemplateSlot{suit} suit.}
\CardsTemplateRow{Presented in full view is the \CardsTemplateSlot{number} of \CardsTemplateSlot{suit}.}{In full view, the suit is \CardsTemplateSlot{suit}.}
\CardsTemplateRow{The photograph identifies the \CardsTemplateSlot{number} of \CardsTemplateSlot{suit}.}{The photograph identifies the suit as \CardsTemplateSlot{suit}.}
\CardsTemplateRow{The main subject is the \CardsTemplateSlot{number} of \CardsTemplateSlot{suit}.}{The main suit is \CardsTemplateSlot{suit}.}
\CardsTemplateRow{A well-defined \CardsTemplateSlot{number} of \CardsTemplateSlot{suit} appears in the image.}{The well-defined suit is \CardsTemplateSlot{suit}.}
\CardsTemplateRow{The card is a carefully illustrated \CardsTemplateSlot{number} of \CardsTemplateSlot{suit}.}{The emblem belongs to \CardsTemplateSlot{suit}.}
\CardsTemplateRow{A clear corner index identifies the \CardsTemplateSlot{number} of \CardsTemplateSlot{suit}.}{The corner emblem shows \CardsTemplateSlot{suit}.}
\CardsTemplateRow{The central design belongs to the \CardsTemplateSlot{number} of \CardsTemplateSlot{suit}.}{The surrounding emblems indicate \CardsTemplateSlot{suit}.}
\CardsTemplateRow{The image captures a fully visible \CardsTemplateSlot{number} of \CardsTemplateSlot{suit}.}{The clearly printed suit is \CardsTemplateSlot{suit}.}
\CardsTemplateRow{The card face belongs to the \CardsTemplateSlot{number} of \CardsTemplateSlot{suit}.}{The face uses \CardsTemplateSlot{suit} emblems.}
\CardsTemplateRow{This standard playing card is the \CardsTemplateSlot{number} of \CardsTemplateSlot{suit}.}{The standard suit is \CardsTemplateSlot{suit}.}
\CardsTemplateRow{The photograph provides an unobstructed view of the \CardsTemplateSlot{number} of \CardsTemplateSlot{suit}.}{The unobstructed suit is \CardsTemplateSlot{suit}.}
\CardsTemplateRow{The displayed card combines the value \CardsTemplateSlot{number} with the \CardsTemplateSlot{suit} suit.}{The displayed suit is \CardsTemplateSlot{suit}.}
\CardsTemplateRow{A sharply defined \CardsTemplateSlot{number} of \CardsTemplateSlot{suit} occupies the frame.}{The sharply defined suit is \CardsTemplateSlot{suit}.}
\CardsTemplateRow{The card can be read as the \CardsTemplateSlot{number} of \CardsTemplateSlot{suit}.}{Its readable suit is \CardsTemplateSlot{suit}.}
\CardsTemplateRow{The visible corner and center markings reveal the \CardsTemplateSlot{number} of \CardsTemplateSlot{suit}.}{The suit markings reveal \CardsTemplateSlot{suit}.}
\CardsTemplateRow{The card shown in detail is the \CardsTemplateSlot{number} of \CardsTemplateSlot{suit}.}{The detailed suit is \CardsTemplateSlot{suit}.}
\CardsTemplateRow{The image centers on the \CardsTemplateSlot{number} of \CardsTemplateSlot{suit}.}{The centered suit is \CardsTemplateSlot{suit}.}
\CardsTemplateRow{A conventional deck card appears as the \CardsTemplateSlot{number} of \CardsTemplateSlot{suit}.}{The card suit is \CardsTemplateSlot{suit}.}
\CardsTemplateRow{The subject is unmistakably the \CardsTemplateSlot{number} of \CardsTemplateSlot{suit}.}{The unmistakable suit is \CardsTemplateSlot{suit}.}
\CardsTemplateRow{Both the corner index and central artwork identify the \CardsTemplateSlot{number} of \CardsTemplateSlot{suit}.}{The corner emblem indicates \CardsTemplateSlot{suit}.}
\CardsTemplateRow{The card is shown face up as the \CardsTemplateSlot{number} of \CardsTemplateSlot{suit}.}{The face-up suit is \CardsTemplateSlot{suit}.}
\CardsTemplateRow{The image gives a direct view of the \CardsTemplateSlot{number} of \CardsTemplateSlot{suit}.}{The directly visible suit is \CardsTemplateSlot{suit}.}
\CardsTemplateRow{The playing card carries the designation \CardsTemplateSlot{number} of \CardsTemplateSlot{suit}.}{Its suit designation is \CardsTemplateSlot{suit}.}
\CardsTemplateRow{A balanced composition displays the \CardsTemplateSlot{number} of \CardsTemplateSlot{suit}.}{The suit shown in the design is \CardsTemplateSlot{suit}.}
\CardsTemplateRow{The card face clearly communicates the \CardsTemplateSlot{number} of \CardsTemplateSlot{suit}.}{The communicated suit is \CardsTemplateSlot{suit}.}
\CardsTemplateRow{The photographed card can be identified as the \CardsTemplateSlot{number} of \CardsTemplateSlot{suit}.}{The identified suit is \CardsTemplateSlot{suit}.}
\CardsTemplateRow{The visual details establish this as the \CardsTemplateSlot{number} of \CardsTemplateSlot{suit}.}{The visual details establish the suit as \CardsTemplateSlot{suit}.}
\CardsTemplateRow{A complete view of the card reveals the \CardsTemplateSlot{number} of \CardsTemplateSlot{suit}.}{The complete view reveals the suit \CardsTemplateSlot{suit}.}
\CardsTemplateRow{The card's printed elements identify the \CardsTemplateSlot{number} of \CardsTemplateSlot{suit}.}{The printed suit marks are \CardsTemplateSlot{suit}.}
\CardsTemplateRow{The image records the \CardsTemplateSlot{number} of \CardsTemplateSlot{suit} in full detail.}{The recorded suit is \CardsTemplateSlot{suit}.}
\CardsTemplateRow{The card's identity is the \CardsTemplateSlot{number} of \CardsTemplateSlot{suit}.}{The suit component of the identity is \CardsTemplateSlot{suit}.}
\CardsTemplateRow{The photographed subject is a \CardsTemplateSlot{number} of \CardsTemplateSlot{suit}.}{The clear suit is \CardsTemplateSlot{suit}.}
\CardsTemplateRow{Every visible marking points to the \CardsTemplateSlot{number} of \CardsTemplateSlot{suit}.}{The suit markings point to \CardsTemplateSlot{suit}.}
\CardsTemplateRow{The image offers a precise view of the \CardsTemplateSlot{number} of \CardsTemplateSlot{suit}.}{The precise suit is \CardsTemplateSlot{suit}.}
\nobreak
\hrule height \heavyrulewidth
\endgroup

\subsubsection{Example Full-Target Pair}
\noindent\textbf{Full description.}\enspace
The image presents the ace of hearts.\par
\noindent\textbf{Target description.}\enspace
Its suit is hearts.\par

\section{Variability Across Random Seeds}
\label{sec:supp-seed-variability}

To assess how consistently CoTeMIC performs across runs, we evaluate each clustering criterion with 10 random seeds. Table~\ref{tab:supp-main-seed-results} reports the mean and standard deviation of normalized mutual information (NMI), clustering accuracy (ACC), and Rand index (RI) for each dataset and criterion. The mean summarizes performance across runs, and the standard deviation quantifies run-to-run variability. The results indicate that CoTeMIC performs consistently across random seeds for most clustering criteria.

\end{document}